\documentclass{article}

\usepackage{microtype}
\usepackage{graphicx}
\usepackage{subcaption}
\usepackage{booktabs} 

\usepackage{hyperref}

\usepackage[accepted]{icml2026}

\usepackage{amsmath}
\usepackage{amssymb}
\usepackage{mathtools}
\usepackage{amsthm}

\usepackage[capitalize,noabbrev]{cleveref}
\usepackage{listings}
\usepackage[most]{tcolorbox}
\usepackage{multirow}
\usepackage{float}

\theoremstyle{plain}

\theoremstyle{definition}

\theoremstyle{remark}

\usepackage[textsize=tiny]{todonotes}

\icmltitlerunning{Clarify Before You Draw: Proactive Agents for Robust Text-to-CAD Generation}

\begin{document}

\twocolumn[
  \icmltitle{Clarify Before You Draw: Proactive Agents for Robust Text-to-CAD Generation}



  \icmlsetsymbol{equal}{*}

  \begin{icmlauthorlist}
    \icmlauthor{Bo Yuan}{Gatech}
    \icmlauthor{Zelin Zhao}{Gatech}
    \icmlauthor{Petr Molodyk}{Gatech}
    \icmlauthor{Bin Hu}{UIUC}
    \icmlauthor{Yongxin Chen}{Gatech}
  \end{icmlauthorlist}

  \icmlaffiliation{Gatech}{Georgia Institute of Technology}
  \icmlaffiliation{UIUC}{University of Illinois Urbana-Champaign}

  \icmlcorrespondingauthor{Yongxin Chen}{yongchen@gatech.edu}

  \icmlkeywords{Machine Learning, ICML}

  \vskip 0.3in
]



\printAffiliationsAndNotice{}  

\begin{abstract}
Large language models have recently enabled text-to-CAD systems that synthesize parametric CAD programs (e.g., CadQuery) from natural-language prompts. In practice, however, geometric descriptions can be under-specified or internally inconsistent: critical dimensions may be missing and constraints may conflict. However, existing fine-tuned models tend to reactively follow the user’s instructions and hallucinate  dimensions when the text is ambiguous. To address this, we propose a proactive agentic framework for text-to-CadQuery generation, named as \textbf{ProCAD}, that resolves specification issues before code synthesis. Our framework pairs a proactive clarifying agent, which audits the prompt and asks targeted clarification questions only when necessary to produce a self-consistent specification, with a CAD coding agent that translates the specification into an executable CadQuery program. We fine-tune the coding agent based on a curated high-quality text-to-CadQuery dataset and train the clarifying agent via agentic SFT on clarification trajectories. Experiments show that proactive clarification significantly improves robustness to ambiguous prompts while keeping interaction overhead low. ProCAD outperforms frontier closed-source models, including Claude Sonnet 4.5, reducing the mean Chamfer distance by 79.9\% and lowering the invalidity ratio from 4.8\% to 0.9\%. Our code and datasets are made publicly available on \url{https://github.com/BoYuanVisionary/Pro-CAD}.
\end{abstract}
\section{Introduction}

Computer-Aided Design (CAD) is central to modern engineering and manufacturing, enabling precise, editable 3D models that support downstream simulation and fabrication~\citep{briere2012comparing}. Yet creating CAD models remains labor-intensive and expertise-heavy, making rapid iteration costly and limiting accessibility~\citep{robertson2002cad}. Recently, the usage of text-to-CAD methods has gained popularity as they use natural language as an intuitive interface for CAD model creation, potentially lowering the expertise barrier and enabling faster iteration by translating user descriptions into parametric CAD models~\citep{badagabettu2024query2cad,li2024cad,khan2024text2cad}. In most existing CAD generation methods, CAD models are represented either as parametric command sequences or as B-rep representations~\citep{wu2021deepcad,xu2024cad, khan2024cad}.

With the advances in large language models (LLMs) and vison language models (VLMs) for language understanding and program synthesis~\citep{chen2021evaluating,austin2021program,jiang2024survey}, language models have also been applied to text-to-CAD to translate natural language instructions into structured CAD code~\citep{xie2025text,guan2025cad,kolodiazhnyi2025cadrille,jia2025meml}. Among various CAD code representations, CadQuery~\citep{cadquery_2_4_0_2024}, a Python-based parametric scripting language, has been increasingly adopted in recent work~\citep{niu2025intent, kolodiazhnyi2025cadrille, xie2025text, guan2025cad}, mostly because modern LLMs tend to be particularly effective at generating Python code~\citep{qing2025effibench}. Recent CadQuery-based methods have also demonstrated strong performance on text-to-CAD generation benchmarks~\citep{kolodiazhnyi2025cadrille,guan2025cad}. In this work, we represent CAD models as executable CadQuery programs, following the line of prior work.

The majority of prior works focuses on training LLMs or VLMs to generate CadQuery code, typically assuming that the input text can precisely specify the target shape. However, in practice, natural-language geometric descriptions are often under-specified or internally inconsistent: critical dimensions may be omitted and constraints may conflict, so generation can fail even when the intended shape is simple~\citep{becattini2013introduction, cheong2014investigating}. Prior text-to-CAD methods largely follow two paradigms—one-shot generation via fine-tuned LLMs~\citep{zhang2024flexcad,li2025cad} and feedback-based refinement using parametric or visual signals~\citep{alrashedy2024generating,li2025seek}. Both paradigms typically treat the user prompt as a reliable specification, implicitly requiring users to provide precise and consistent geometric constraints. 

This issue motivates a shift from static text-to-CAD generation with a single LLM to dynamic, proactive clarification with an agentic system. Proactive agents not only follow the user’s request to improve generation success, but also identify missing or conflicting information and ask only essential clarification questions to minimize interruption and frustration~\citep{lu2024proactive,sun2025training}. To mitigate the ambiguity in text descriptions, we design a two-agent system consisting of a proactive clarifying agent and a CAD code generation agent. The clarifying agent first audits the user prompt, interacts with the user to resolve missing and conflicting dimensions, and then produces a complete, self-consistent text specification, which is finally passed to the coding agent to generate the CAD program. 


\begin{figure*}[th!]
  \begin{center}
\centerline{\includegraphics[width=0.85\textwidth]{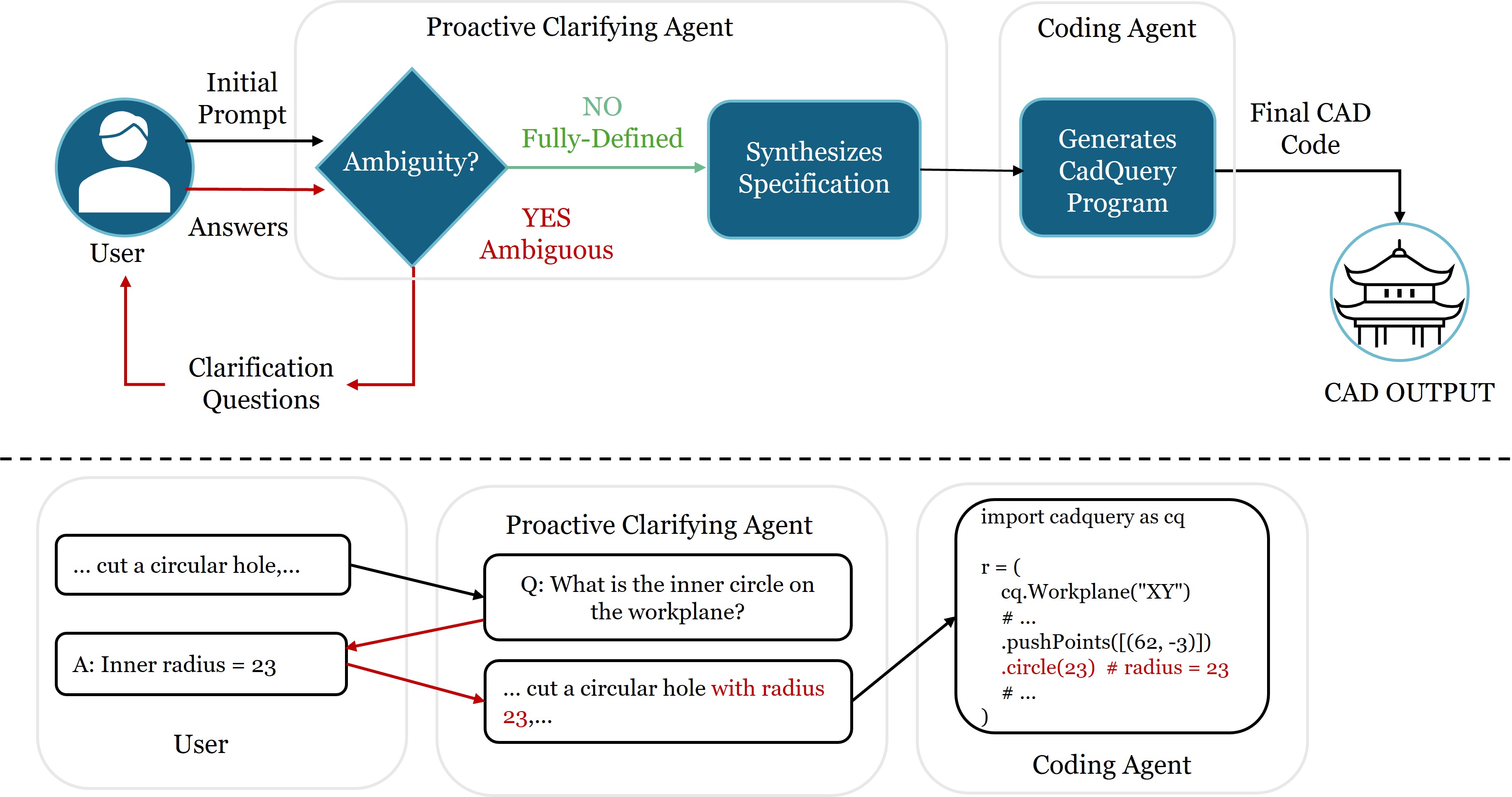}} 
\vspace{-1mm}
    \caption{
     Diagram of our two-agent text-to-CadQuery pipeline. A proactive clarifying agent audits the user prompt, asks targeted clarification questions when needed, and outputs a standardized specification, which a coding agent then outputs a CadQuery program. We also provide one example where the ambiguous prompt does not include the radius of the inner circle.
    }
    \label{fig:pipeline}
  \end{center}
\vspace{-5mm}
\end{figure*}

To build this proactive agentic system, we fine-tune open-source models on a curated, high-quality text-to-CadQuery dataset of 10K unambiguous samples. We develop a new data creation pipeline that generates human-like, concise natural-language specifications for CAD models from DeepCAD~\citep{wu2021deepcad}. Unlike Text2CAD~\citep{khan2024text2cad}, which is built from a minimal JSON representation, our approach instead uses CadQuery code as the canonical representation. We also apply LLM-based verification and human checks to remove potential data leakage and ensure consistency between the natural language description and the intended geometry. The resulting high-quality text-to-CadQuery dataset is used to fine-tune the CAD coding agent. On top of this, we construct a synthetic dataset to simulate ambiguous text descriptions by perturbing the verified specifications to induce syntactic ambiguity, while keeping the corresponding corrected specifications as targets; this dataset is used to supervise the proactive agent to ask minimal  questions and produce a finalized specification.

We summarize our main contributions as follows:
\begin{enumerate}
    \item We propose a shift from static one-shot generation or post-hoc refinement to a dynamic proactive agentic framework for text-to-CadQuery, where a clarifying agent detects missing or conflicting constraints and asks minimal clarification questions, and a coding agent then synthesizes executable CadQuery code from the resulting specification.
    \item We fine-tune our coding agent, \emph{ProCAD-coder}, using only 1.6K carefully curated samples, yet achieve superior performance on unambiguous text-to-CadQuery generation. These 1.6K samples are drawn from a new data-creation pipeline that produces a curated 10K text-to-CadQuery dataset, where specifications are screened with LLM-based checks and  human verifications.
    \item We fine-tune our clarifying agent, \emph{ProCAD-clarifier}, via agentic SFT on a synthetic dataset of 6{,}063 samples containing full agentic trajectories to resolve ambiguous prompts. The resulting system outperforms frontier models, including Claude Sonnet 4.5 and GPT-4o-mini, on communication cost, corrected-prompt quality, and downstream geometry quality.

\end{enumerate}

\section{Related Work}

\paragraph{Text-to-CAD and Parametric Generation.}
Early CAD generation relied on static representations like voxels or meshes~\citep{wu2021deepcad}, while recent approaches focus on parametric, editable command sequences~\citep{khan2024cad}. Text-to-CAD has emerged to lower entry barriers~\citep{badagabettu2024query2cad,li2024cad,khan2024text2cad} and typically treats generation as a direct translation problem. Consequently, current methods can struggle with ambiguous, incomplete, or inconsistent prompts~\citep{becattini2013introduction}.
\vspace{-1mm}
\paragraph{LLMs for CAD Generation.}
Recent works leverage LLMs to generate structured CAD scripts~\citep{xie2025text,guan2025cad,kolodiazhnyi2025cadrille,jia2025meml}, with CadQuery gaining traction due to its Python-based syntax~\citep{cadquery_2_4_0_2024,niu2025intent,qing2025effibench}. Unlike existing one-shot systems~\citep{kolodiazhnyi2025cadrille,guan2025cad} or those relying on post-hoc execution feedback~\citep{alrashedy2024generating,li2025seek,wang2025text}, our work introduces a method to audit and correct specification errors \emph{before} code generation, preventing downstream failures.
\vspace{-1mm}
\paragraph{Text-to-CadQuery Datasets.}
High-quality datasets pairing expert descriptions with executable CadQuery code are scarce. Existing resources like LLM4CAD~\citep{li2024llm4cad} and Query4CAD~\citep{badagabettu2024query2cad} are often small or limited in scope. Many studies rely on expert-level descriptions in Text2CAD~\citep{xie2025text,guan2025cad,khan2024text2cad}, but these are frequently verbose, noisy, or contain misleading scaling operations~\citep{govindarajan2025cadmium} (see Appendix~\ref{app:failure_mode}). Previous attempts to pair these descriptions with CadQuery code~\citep{kolodiazhnyi2025cadrille,rukhovich2025cad} often overlook critical discrepancies in units and commands. \noindent {We provide a more detailed discussion of related works in Appendix~\ref{app:related_work}.}

\section{Proactive Agentic System}

In this work, we study \emph{text-to-CadQuery} generation, where a natural-language specification $p$ is translated into an executable CadQuery program $y$. We allow the specification $p$ to be ambiguous, yet aim to recover the correct CadQuery code with minimal user interruption: if $p$ is fully specified, we directly generate $y$; otherwise, we proactively ask for clarification from users before generating $y$.



To this end, instead of relying on a single model to resolve an ambiguous prompt end-to-end, we decompose the task into three explicit stages: (1) detecting ambiguous aspects of the description and asking targeted clarification questions; (2) incorporating the user's feedback to produce a corrected text description; and (3) generating the final CadQuery program from the corrected description. This decomposition makes the process more controllable and interpretable and fits the high standards of engineering design. To implement this paradigm, we propose a two-agent system consisting of a proactive clarification agent and a CAD coding agent, as shown in Figure~\ref{fig:pipeline}. 

More formally, we model the proactive clarification agent $\pi_{\phi}$ as a finite-horizon Markov decision process $\mathcal{M} = \left(\mathcal{S}, \mathcal{A}, R \right)$ where the environment corresponds to the user. We omit the transition kernel for brevity. The interaction starts from the original user prompt $p$. A state $s \in \mathcal{S}$ captures the current context,
$
s = (p, h),
$
where $h$ is the conversation history consisting of previously asked questions and user answers (with $h=\emptyset$ at the start). At each round, the proactive agent $\pi_{\phi}$ either accepts the current specification or asks a clarification question:
\begin{equation}
a \in \mathcal{A} = \{\mathrm{ACCEPT}\} \cup \{\mathrm{ASK}(u): u \in \mathcal{U}\},
\end{equation}
where $\mathcal{U}$ denotes the space of natural-language questions. If $a=\mathrm{ASK}(u)$, the user provides an answer $v$ and the history is updated as $h \leftarrow h \cup \{(u,v)\}$. When the agent chooses $\mathrm{ACCEPT}$, it outputs a finalized, self-consistent specification $\hat{p}$ based on $(p,h)$, which is then passed to the coding agent $\pi_{\theta}$ to generate the CadQuery program $y$.

The two-agent system should maximize the reward that captures both the geometric fidelity of the model and communication overhead with the user. Let $\mathrm{CD}(y)$ denote the Chamfer distance between the generated mesh from code $y$ and the ground-truth mesh, and let $C(h)$ be a nonnegative cost that measures interaction burden, such as the number of rounds, total token length, or latency. We define the reward of the clarifying agent as
$
R = -\mathrm{CD}(y) -\lambda\, C(h),
$
where $\lambda \ge 0$ controls the trade-off between reconstruction accuracy and interaction cost. The objective is to learn policies $\pi_{\phi}$ and $\pi_{\theta}$ that maximize the expected return, equivalently minimizing $\mathrm{CD}$ while keeping $C(h)$ small. For each step, we carefully design system prompts to enforce format completeness and clearly specify each agent’s role. Full prompts are provided in Appendix~\ref{app:system_prompts}.

Our agentic system serves as a flexible framework that can be instantiated with different combinations of a clarifying agent and a CAD coding agent, using either commercial or open-source models. To further improve performance, we design a two-stage training process for both agents using a carefully curated text-to-CadQuery dataset that includes both unambiguous and ambiguous text prompts (Section~\ref{section:data_collection}). First, we fine-tune the coding agent on our high-quality text-to-CadQuery dataset of unambiguous text descriptions (Section~\ref{section:coding_agent_traning}). Second, we generate synthetic expert agentic trajectories for resolving ambiguity in natural-language descriptions and use them to fine-tune the clarification agent via agentic supervised fine-tuning (Section~\ref{section:clarifying_agent_method}). Our agentic system, ProCAD, pairs a fine-tuned Qwen2.5-7B-Instruct model (ProCAD-clarifier) as the clarifying agent with another fine-tuned Qwen2.5-7B-Instruct model (ProCAD-coder) as the coding agent, and outperforms even frontier coding models, Claude Sonnet 4.5~\citep{anthropic2025claude-sonnet-4-5} in both communication cost and geometric fidelity (Section~\ref{section:experiment}).

\section{Data Annotation pipeline for high-quality text-to-CadQuery dataset}\label{section:data_collection} 


In contrast to prior pipelines (as discussed in Section~\ref{section:related_work_dataset}) that generate CadQuery code from text descriptions, we instead start from the raw CadQuery programs and generate precise, high-quality text descriptions. CadQuery is a highly interpretable programming language that typically uses common CAD operations to construct geometry. As a result, it contains the complete information needed to reconstruct an accurate and detailed textual specification.
We observe that~\citet{rukhovich2025cad} reconstructs CadQuery programs for DeepCAD models from point clouds and builds a CadQuery dataset of approximately 17K samples. Building on this, we render the shape from four different viewpoints and prompt a strong vision-language model, GPT-5-mini~\citep{singh2025openai}, with both the images and the CadQuery program to produce the corresponding text. 




We first apply standard deduplication procedures~\citep{xu2022skexgen, xu2023hierarchical} to the original DeepCAD shapes. We only keep the subset of deduplicated shapes for which CadQuery programs are available from~\citep{rukhovich2025cad}. Finally, we filter out samples whose generated geometry deviates from the reference shape by more than a preset Chamfer distance threshold. This step removes only a small fraction of samples, indicating that the adopted CadQuery corpus is generally of high quality. See Table~\ref{tab:cd_stats} for details.

\begin{figure*}[ht]
  \begin{center}
\centerline{\includegraphics[width=0.85\textwidth]{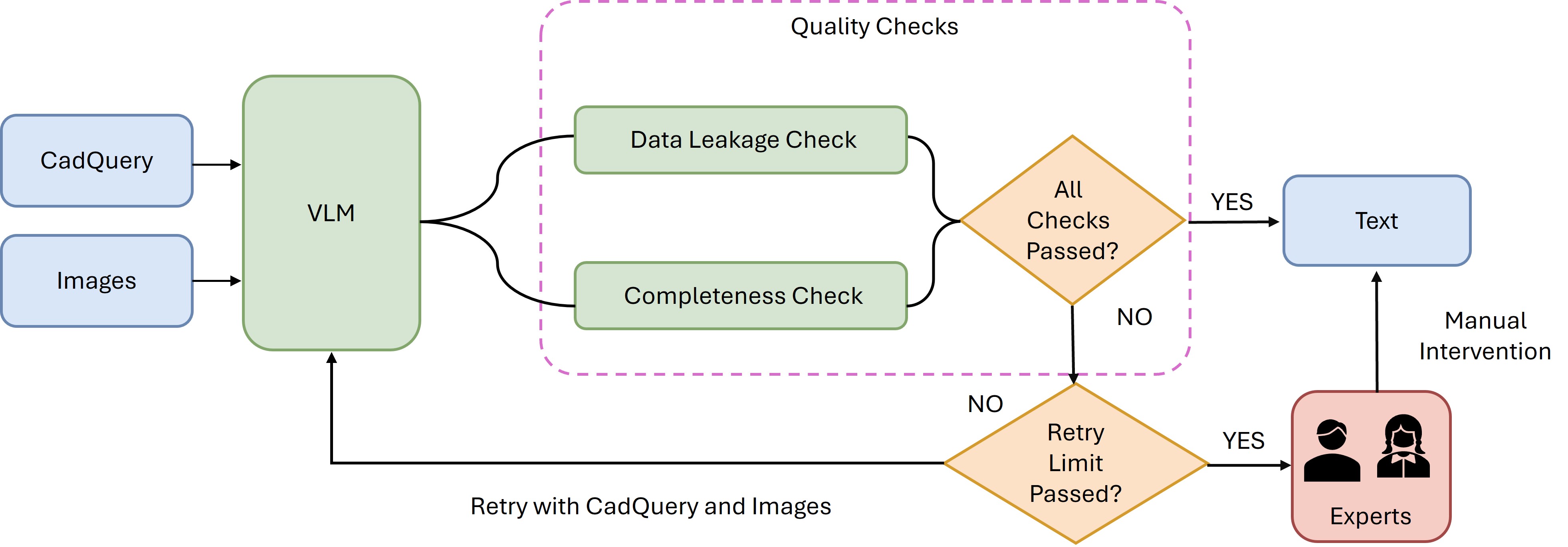}} 
    \caption{
     Semi-automatic annotation pipeline: render each shape and use its CadQuery to prompt a frontier VLM for a text description; accept only if it passes (i) code-leakage filtering and (ii) completeness verification by regenerating CadQuery with Chamfer distance below a threshold; otherwise retry up to a limit. If the maximum number of retries is reached, we defer the case to CAD experts for manual review.
    }
    \label{fig:anotation_pipeline}
  \end{center}
\vspace{-10mm}
\end{figure*}


At the same time, using CadQuery as an input introduces a new challenge for building text-to-CadQuery data: the generated description may leak code snippets from the source program. To mitigate this risk, our system prompt explicitly instructs the model to produce natural descriptions \emph{without} reproducing any CadQuery surface form. The system prompt is provided in Appendix~\ref{app:text_generation_prompt}. In addition, we adopt a generate-then-verify pipeline~\citep{madaan2023self} to detect and filter potential leakage: each generated text will have one data leakage check. This LLM-based check is designed to catch raw code or near-verbatim fragments, while avoiding overly strict false positives---for example, it does \emph{not} flag ordinary geometric terms, e.g., origin, workplane or coordinate tuples as leakage in isolation. Details of the leakage-check prompt are also provided in Appendix~\ref{app:text_generation_prompt}.

To further ensure that the generated natural language description is complete and unambiguous, we add an LLM based completeness check. Concretely, we provide the generated description alone to GPT-5-mini without the original CadQuery program and prompt it to synthesize CadQuery code. We then execute the generated program to obtain a three dimensional mesh and compute the Chamfer distance to the ground truth mesh. 


Note that this completeness check can be overly strict: even prompts with fully specified descriptions may fail due to the model’s limitations in generating correct CadQuery code. Motivated by inference time scaling laws \citep{wu2025inference,snell2025scaling}, where models often succeed given multiple independent attempts, we incorporate a simple and efficient retry mechanism. If a sample fails either the leakage check or the completeness check, we rerun the entire generation and validation loop up to three times, each time sampling a new description and reapplying both checks.

If all three attempts fail, we escalate this CAD model to a team of CAD experts for manual review and supervision. This design balances the trade off between human effort and automated evaluation during data generation. Empirically, more than 80\% of samples pass both checks with retry, which substantially reduces the amount of manual intervention required. See Figure~\ref{fig:anotation_pipeline} for the complete pipeline. 


Compared with the Text2CAD pipeline~\citep{khan2024text2cad}, our approach offers three key advantages. First, we jointly condition the description generator on \emph{both} the multi-view renderings and the CadQuery program, whereas Text2CAD uses visual and symbolic signals separately; this joint conditioning yields descriptions that are better grounded in the underlying geometry and less likely to omit critical constraints~\citep{ngiam2011multimodal}. Second, we rely on a frontier VLM, GPT-5-mini, which we find produces substantially more consistent and higher-quality descriptions than smaller models, reducing noise in the resulting supervision. Third, we incorporate two checks for data leakage and completeness with a retry mechanism to further improve quality. To avoid biasing the dataset toward overly simple samples, we additionally route cases that repeatedly fail these checks to human experts.

\section{ProCAD Training}

\subsection{Coding Agent Training}\label{section:coding_agent_traning}
To train the coding agent, we perform standard supervised fine-tuning (SFT) of an open-source pretrained language model on a paired dataset $\mathcal{D}=\{(p, y)\}$, where $p$ is an unambiguous prompt obtained in Secion~\ref{section:data_collection} and $y$ is the corresponding CadQuery program released by~\citet{rukhovich2025cad}. Let $P_{\theta}(\cdot \mid p_0, p)$ denote the causal language model distribution over CadQuery programs conditioned on the system prompt $p_0$ and the input prompt $p$. Here $p_0$ specifies the required output format and coding conventions that the generated CadQuery program must follow. We minimize the negative log-likelihood objective
\begin{equation}
\min_{\theta}\; \mathcal{L}_{}(\theta)
= \mathbb{E}_{(p,y)\sim \mathcal{D}}
\left[-\log \pi_{\theta}(y \mid p_0, p)\right].
\end{equation}

\subsection{Proactive Clarifying Agent Training}\label{section:clarifying_agent_method}

Real user inputs are often noisy and may be under specified or contain intrinsically contradictory dimensions. Training a coding agent with standard SFT alone is therefore insufficient to reliably detect and resolve such specification errors. A natural mitigation is to incorporate richer feedback~\citep{alrashedy2024generating, li2025seek, an2026pr}, such as images or point clouds of the rendered models as targets, but these signals are often too coarse to enforce precise metric constraints and may not penalize small yet consequential deviations. We argue that this mismatch is especially critical in CAD design, where even minor dimensional errors can violate product requirements and cause downstream manufacturing failures.

In practice, fully specifying every dimension of a complex CAD part can be difficult for users, whereas providing a few dimensions in response to targeted questions is often easier. Hence in our experiments, we assume that the user can provide correct answers to any asked question as long as the question itself is clear. Under this assumption, an optimal policy should minimize the number of interaction rounds, since additional rounds only increase communication cost and lengthen the context presented to the agent. Consequently, the general multi-round optimization reduces to a two-round policy: in the first round, the agent either directly accepts the original prompt and output the corrected text $\hat{p}=p$, or asks a set of targeted clarification questions $\{u_j\}$ in a single message. The question should be clear and specific enough that users can give right dimensions. In the second round, after receiving the user answers, the agent deterministically accepts and outputs the corrected specification $\hat{p}$, which is then passed to the coding agent.


We train the clarifying agent from two kinds of supervision, and in both cases the expert target is a JSON-formatted output. For unambiguous prompts, the dataset is
\(
\mathcal{D}_{\mathrm{acc}}=\{p^{(i)}, y^{(i)}_{\mathrm{acc}}\},
\)
where the target JSON is
$
y^{(i)}_{\mathrm{acc}}
=
\{\text{is misleading}:\text{False},\;
  \text{standardized prompt}:p^{(i)}\}.
$
For ambiguous prompts, we store clarification trajectories in the form
$
(\hat{p}^{(j)},\mathbf{q}^{(j)}, \mathbf{a}^{(j)},  p^{(j)})
$,
where \(\mathbf{q}^{(j)}\) are the clarification questions and \(\mathbf{a}^{(j)}\) are the corresponding user answers. Supervision is provided in two JSON outputs. The first JSON supervises question generation,
$
y^{(j)}_{\mathrm{ask}}
=
\{\text{is misleading}:\text{True},\;
  \text{questions}:\mathbf{q}^{(j)}\},
$
and the second JSON supervises the final corrected specification,
$
y^{(j)}_{\mathrm{clr}}
=
\{\text{is misleading}:\text{True},\;
  \text{standardized prompt}:{p}^{(j)}\}.
$

We train the model $\pi_{\phi}$ to reproduce these JSON outputs via maximum likelihood. The overall objective sums the three losses where $s_{\phi}$ is the system prompt, as shown in Appendix~\ref{prompt: clarification}.
\begin{align}
\mathcal{L}(\phi)
&=
\mathbb{E}_{(p,y_{\mathrm{acc}})\sim \mathcal{D}_{\mathrm{acc}}}
\Big[-\log \pi_{\phi}\big(y_{\mathrm{acc}} \mid s_{\phi}, p\big)\Big] \nonumber\\
&+
\mathbb{E}_{(\hat{p},y_{\mathrm{ask}})\sim \mathcal{D}_{\mathrm{clr}}}
\Big[-\log \pi_{\phi}\big(y_{\mathrm{ask}} \mid s_{\phi}, \hat{p}\big)\Big] \nonumber\\
&+
\mathbb{E}_{(\hat{p},\mathbf{q},\mathbf{a},y_{\mathrm{clr}})\sim \mathcal{D}_{\mathrm{clr}}}
\Big[-\log \pi_{\phi}\big(y_{\mathrm{clr}} \mid s_{\phi}, \hat{p}, \mathbf{q}, \mathbf{a}\big)\Big].
\end{align}

\section{Experiments}\label{section:experiment}

\paragraph{Metrics} In our experiments, we report two primary metrics. \textbf{Chamfer distance} (CD) measures geometric fidelity between the generated and reference shapes. \textbf{Invalidity Ratio} (IR) is the percentage of generated samples that cannot be executed or rendered into valid CAD objects. Both follow common practice in previous works~\citep{guan2025cad, kolodiazhnyi2025cadrille, xie2025text, wang2025text}. In addition, we use GPT-5-mini as a judge to assess the quality of our text-to-CadQuery dataset and the effectiveness of the resulting user interactions.

\subsection{Text-to-CadQuery dataset}
 \paragraph{Dataset Creation} To construct our high-quality Text-to-CadQuery dataset, we start from the CadQuery dataset of~\citet{rukhovich2025cad} and retain only samples whose reconstructed geometry matches the reference with CD below $2\times 10^{-4}$. The full CD distribution is reported in Table~\ref{tab:cd_stats}.
 We use the same threshold in the completeness check in the data annotation pipeline, where we require that, given only the generated natural-language description, GPT-5-mini can synthesize CadQuery code whose executed geometry attains CD $< 2\times 10^{-4}$. For each sample, we allow up to three retries in the generation-and-verification loop, and find that over $80\%$ of samples pass both the leakage and completeness checks without human intervention. The system prompts used in our annotation pipeline are provided in Appendix~\ref{app:text_generation_prompt}. Each generated description follows a fixed structure with three parts: {General shape}, {Setup}, and {Build description}. {General shape} briefly names the part and its main features, {Setup} specifies the workplane and its origin including any coordinate transforms, and Build description provides step-by-step instructions.

\paragraph{Comparison against Text2CAD}
Table~\ref{tab:prompt_stats_and_pref} compares our dataset with Text2CAD. First, our prompts are substantially shorter, primarily because we retain only the information necessary to reconstruct the CadQuery program, whereas Text2CAD descriptions are designed for command-sequence generation and often include redundant details. Second, we use an LLM-as-judge to assess \emph{clarity} and \emph{human-likeness}. We randomly sample 1{,}000 pairs and ask the judge to choose which description is better under each criterion. Because LLM judges can exhibit position bias~\citep{zheng2023judging, shi2025judging}, where preferences depend on whether an option appears first or second, we evaluate both presentation orders and report results for ``Ours first'' and ``Text2CAD first.'' Across both orders, our prompts are consistently preferred in terms of human-likeness. For clarity, Text2CAD descriptions typically include redundant details, e.g., Euler angles, scaling factors, and translation vectors, that can distract from the core geometric specification, whereas our prompts are more concise and less confusing, leading to higher clarity win rates as well. Finally, it is worth noting that LLM-based judges often exhibit a length bias, tending to prefer longer responses~\citep{saito2023verbosity,dubois2024length}. Since our prompts are markedly shorter, these preference-based evaluations may be biased against our dataset, yet we still outperform Text2CAD, underscoring our data quality. Prompts for LLM-as-judge are shown in Appendix~\ref{section: LLM_judge_prompt}.

\paragraph{Zero-shot performance} Moreover, we evaluate the zero-shot performance of both frontier models and open-source models on Text2CAD and our new dataset, as in Table~\ref{tab:dataset_compare_ir_cdl}. For a fair comparison, we consider the subset of Text2CAD that shares the same shape UIDs as our data. We find that Qwen2.5-7B-Instruct~\citep{bai2025qwen2} performs poorly in zero-shot settings on both datasets, with an invalidity ratio of nearly $85\%$. In contrast, Claude Sonnet 4.5~\citep{anthropic2025claude-sonnet-4-5}, a strong frontier coding model, achieves substantially better results: the zero-shot invalidity ratio decreases from approximately $54.5\%$ on Text2CAD to about $13\%$ on our dataset, and it also attains much lower mean and median Chamfer distances. These results partially support our claim that our dataset is easier to follow and contains fewer misleading specifications. We also observe that most failures of Qwen2.5-7B-Instruct in the zero-shot setting are due to CadQuery syntax errors rather than geometric reasoning. In the following experiments we use Qwen2.5-7B-Instruct as the base model and fine-tune it on our dataset to improve code validity and overall generation quality.

\begin{table}[t]
\centering
\caption{Invalidity ratio (IR) and Chamfer distance (CD) on Ours and Text2CAD. CD is reported in units of $\times 10^3$ (lower is better).}
\label{tab:dataset_compare_ir_cdl}
\small
\setlength{\tabcolsep}{1pt} 
\renewcommand{\arraystretch}{0.9} 
\begin{tabular}{@{}llccc@{}} 
\toprule
Model & Dataset & IR (\%)$\downarrow$ & Mean CD$\downarrow$ & Median CD$\downarrow$ \\
\midrule
Claude 4.5 Sonnet & Ours     & \textbf{11.8} & \textbf{2.580} & \textbf{0.074} \\
                  & Text2CAD & 56.6          & 27.244         & 12.549 \\
\midrule
Qwen2.5 7B Instruct & Ours     & \textbf{82.9} & 4.489 & \textbf{0.100} \\
                    & Text2CAD & 86.9          & \textbf{4.434} & 0.103 \\
\bottomrule
\end{tabular}
\vspace{-2mm} 
\end{table}

\subsection{ProCAD-coder}
\paragraph{Setup} We sample 1.6K examples for training and 1K examples for testing from our 10K dataset. Our coding agent is initialized from Qwen2.5-7B-Instruct, which takes a natural-language prompt as input and outputs a standardized CadQuery program. We perform full-parameter fine-tuning on two H200 GPUs with batch size 16, learning rate $10^{-5}$, and two training epochs. This results in 200 total optimization steps and completes in under 10 minutes.

We find that even this lightweight fine-tuning already yields large gains on the test set: the invalidity ratio drops from nearly $86.9\%$ to $0.9\%$, while the median Chamfer distance reaches $6.6\times 10^{-5}$. Notably, this result is competitive with, and in some cases better than, Claude~4.5 Sonnet, which attains an invalidity ratio of about $13\%$ with median Chamfer distance $7.7\times 10^{-5}$. By comparison, prior work typically relies on more than $100$K supervised examples, often combined with additional refinement or reinforcement learning, to achieve similar reliability~\citep{xie2025text,guan2025cad,an2026pr,kolodiazhnyi2025cadrille}. We attribute our strong performance to both our high-quality data creation pipeline and the use of a strong base model. Notably, even when using the same Qwen2.5 backbone for a fair comparison, prior work trains on more than $150$K samples~\citep{guan2025cad} and additionally applies both SFT and reinforcement learning. See Appendix~\ref{app:train_set_up} for comprehensive comparison.


To ensure a fair comparison, we keep the underlying CAD shapes fixed and use the same training and test splits and identical fine-tuning hyperparameters; we vary only the data construction pipeline, which yields different text descriptions and CadQuery program representations for the same shapes. We consider two baselines: \textbf{Text-CAD}, which follows~\citet{kolodiazhnyi2025cadrille} by directly pairing the expert-level Text2CAD prompts with the CadQuery programs from~\citet{rukhovich2025cad}; and \textbf{JSON-Distill} , which uses the open-source dataset of~\citet{xie2025text} where the text is the expert-level Text2CAD description and the CadQuery program is distilled from Gemini~2.0 Flash based on the minimal JSON representation. 

As shown in Table~\ref{tab:testset_results_ir_cd}, our experimental results further demonstrate the importance of text quality in the text-to-CadQuery task. Moreover, our model outperforms Claude Sonnet 4.5 across all evaluation metrics.
While most prior work focuses on improving the model and training procedure on Text2CAD with various techniques, our findings suggest that a key bottleneck also lies in the quality of the original text descriptions. In particular, the code generation model can be strong enough to produce valid CadQuery programs when the input specification is clear and correct. This observation motivates us to consider a more realistic setting in which user prompts may be ambiguous, and to introduce a proactive clarifying agent that detects specification issues and asks targeted questions before code generation.

\begin{table}[t]
\small
\centering
\caption{Performance on the 1K unambiguous prompts. CD is reported in units of $\times 10^3$ (lower is better).}
\label{tab:testset_results_ir_cd}
\begin{tabular}{lccc}
\toprule
Method & IR (\%)$\downarrow$ & Mean CD$\downarrow$ & Median CD$\downarrow$ \\
\midrule
Ours & \textbf{0.9}  & \textbf{0.108} & \textbf{0.066} \\
Text-CAD & 14.5 & 3.054 & 0.097 \\
JSON-Distill & 5.3 & 23.117 & 8.808 \\
Claude 4.5 Sonnet & 12.9 & 1.580 & 0.077 \\
\bottomrule
\end{tabular}
\end{table}

\begin{table}[t]
\centering
\small
\caption{Prompt length statistics and LLM-judge preference win rates (\%) for Ours vs.\ Text2CAD.}
\label{tab:prompt_stats_and_pref}
\begin{tabular}{lcc}
\toprule
Metric & Ours & Text2CAD \\
\midrule
\multicolumn{3}{l}{\textbf{Length statistics} \(\downarrow\)} \\
Mean length   & \textbf{147.8} & 285.4 \\
Median length & \textbf{119.0} & 228.0 \\
\midrule
\multicolumn{3}{l}{\textbf{Win rate \%}, \(\uparrow\) } \\
Clarity, Ours first       & \textbf{98.4} & 1.6 \\
Clarity, Text2CAD first   & \textbf{66.0} & 34.0 \\
Human-likeness, Ours first      & \textbf{100.0} & 0.0 \\
Human-likeness, Text2CAD first  & \textbf{96.5} & 3.5 \\
\bottomrule
\end{tabular}
\end{table}

\subsection{ProCAD-clarifier}
\begin{table*}[!ht]
\centering
\small
\renewcommand{\arraystretch}{1.15} 
\caption{Performance on the test set with 2{,}469 ambiguous prompts, where user responses are simulated by GPT-5-mini. In the two-agent setting (Figure~\ref{fig:pipeline}), we fix the coding agent and vary the clarification agent. Bold and underline denote the best and second-best values.}
\label{tab:agentic_vs_single}
\setlength{\tabcolsep}{4pt}
\begin{tabular}{l l cc ccc}
\hline
\textbf{Setting} & \textbf{Model} 
& \textbf{Efficiency} $\uparrow$
& \textbf{Resolution} $\uparrow$
& \textbf{Mean CD}  $\downarrow$
& \textbf{Median CD}  $\downarrow$
& \textbf{IR \%} $\downarrow$ \\
\hline
\multirow{4}{*}{\textbf{Single-model}}
& Cadrille~\citep{kolodiazhnyi2025cadrille}
& -- & -- & 55.43 & 44.92 & 20.7\% \\
& Qwen 2.5-7B-Instruct 
& -- & -- & 10.94 & 1.03 & 68.2\% \\
& GPT-4o-mini
& -- & -- & 12.58 & 0.78 & 28.7\% \\
& Claude Sonnet 4.5 
& -- & -- & 7.80 & 0.19 & 14.6\% \\
\hline

\multirow{3}{*}{\shortstack[l]{\textbf{Two-agent} \\ \scriptsize (coding=Claude 4.5 Sonnet)}}
& Qwen 2.5-7B-Instruct
& 0.6606 & 0.6487 & 11.56 & 0.33 & 4.5\% \\
& GPT-4o-mini
& 0.5788 & 0.7814 & 9.98 & 0.14 & 7.5\% \\
& Claude Sonnet 4.5
& 0.8255 & \underline{0.9329} & 3.10 & \underline{0.09} & 4.8\% \\
& \textbf{ProCAD-clarifier (Ours)}
& \textbf{0.9665} & 0.9327 & \underline{0.85} & \textbf{0.08} & 4.0\% \\
\hline

\multirow{4}{*}{\shortstack[l]{\textbf{Two-agent} \\ \scriptsize (coding=ProCAD-coder)}}
& Qwen 2.5-7B-Instruct
& 0.6706 & 0.6597 & 11.68 & 0.33 & 4.1\% \\
& GPT-4o-mini 
& 0.5677 & 0.7712 & 9.38 & 0.12 & 3.3\% \\
& Claude Sonnet 4.5
& 0.8485 & 0.9120 & 2.69 & \textbf{0.08} & \underline{2.3\%} \\
& \textbf{ProCAD-clarifier (Ours)} 
& \underline{0.9654} & \textbf{0.9341} & \textbf{0.63} & \textbf{0.08} & \textbf{0.9\%} \\
\hline
\end{tabular}
\vspace{1mm}
\end{table*}

\begin{table*}[!ht]
\centering
\small
\renewcommand{\arraystretch}{1.15} 
\caption{Performance on the test set with ambiguous prompts, where user responses are simulated by {Claude 4.5 Haiku} for out-of-distribution task. In the two-agent setting, we fix the coding agent and vary the clarification agent.}
\label{tab:agentic_vs_single_haiku}
\setlength{\tabcolsep}{4pt}
\begin{tabular}{l l cc ccc}
\hline
\textbf{Setting} & \textbf{Model} 
& \textbf{Efficiency} $\uparrow$
& \textbf{Resolution} $\uparrow$
& \textbf{Mean CD}  $\downarrow$
& \textbf{Median CD}  $\downarrow$
& \textbf{IR \%} $\downarrow$ \\
\hline

\multirow{2}{*}{\shortstack[l]{\textbf{Two-agent} \\ \scriptsize (coding = Claude 4.5 Sonnet)}}
& Claude Sonnet 4.5
& 0.8249 & 0.9367 & 3.06 & {0.09} & 4.2\% \\
& \textbf{ProCAD-clarifier (Ours)}
& \textbf{0.9668} & 0.9354 & \underline{0.63} & \textbf{0.07} & 4.0\% \\
\hline

\multirow{2}{*}{\shortstack[l]{\textbf{Two-agent} \\ \scriptsize (coding = ProCAD-coder)}}
& Claude Sonnet 4.5

& 0.8298 & \underline{0.9372} & 3.14 & \underline{0.08} & \underline{1.7\%} \\
& \textbf{ProCAD-clarifier (Ours)} 
& \underline{0.9658} & \textbf{0.9415} & \textbf{0.46} & \textbf{0.07} & \textbf{0.9\%} \\
\hline
\end{tabular}
\vspace{1mm}
\end{table*}

  \begin{table*}[!ht]
  \centering
  \small
  \renewcommand{\arraystretch}{1.15}
  \caption{Human-centered evaluations. Efficiency and Resolution are scored by human experts. CD is reported in units of $\times 10^3$ (lower
   is better). Bold and underline denote the best and second-best values.}
  \label{tab:human_study}
  \setlength{\tabcolsep}{4pt}
  \begin{tabular}{l l cc ccc}
  \hline
  \textbf{Setting} & \textbf{Model}
  & \textbf{Efficiency} $\uparrow$
  & \textbf{Resolution} $\uparrow$
  & \textbf{Mean CD}  $\downarrow$
  & \textbf{Median CD}  $\downarrow$
  & \textbf{IR \%} $\downarrow$ \\
  \hline

  \multirow{4}{*}{\shortstack[l]{\textbf{Two-agent} \\ \scriptsize (coding = ProCAD-coder)}}
  & Qwen 2.5-7B-Instruct
  & 0.016 & 0.000 & 13.98 & 3.69 & 14.6\% \\
  & GPT-4o-mini
  & 0.251 & 0.287 & 13.81 & 0.14 & 19.5\% \\
  & Claude Sonnet 4.5
  & \underline{0.598} & \underline{0.700} & \underline{9.72} & \underline{0.11} & \textbf{12.2\%} \\
  & \textbf{ProCAD-clarifier (Ours)}
  & \textbf{0.760} & \textbf{0.787} & \textbf{1.28} & \textbf{0.09} & \textbf{12.2\%} \\
  \hline
  \end{tabular}
  \vspace{1mm}
  \end{table*}

After training ProCAD-coder on our new Text2CadQuery dataset, we build the full agentic system. Specifically, to build ProCAD-clarifier, we fine-tune Qwen2.5-7B-Instruct using agentic SFT, as described in Section~\ref{section:clarifying_agent_method}. 


Modeling real user behavior for proactive clarification is extremely challenging: it typically requires large-scale interactive data collection and can exhibit substantial variability across annotators~\citep{testoni2024asking, ito2025enhancing, sahay2025ask}. As the first work toward studying clarification for ambiguous CAD prompts, we therefore adopt a scalable alternative and use GPT-5-mini as a user simulator, following prior work that leverages LLM-based user simulation in dialogue systems~\citep{sekulic2024reliable} and recommendation settings~\citep{zhang2025llm}. This enables us to synthesize ambiguous prompts with predefined ambiguity types in a controllable and reproducible way, while avoiding the cost and noise of large-scale human interaction data.

To generate ambiguous prompts, we prompt GPT-5-mini with a detailed system instruction to perturb the original, verified specifications from our Text-to-CadQuery dataset. See system prompts in Appendix~\ref{app:mislead_prompt_system_prompt}. We focus on two common error types in CAD practice: (i) \emph{under-specified} prompts that omit key dimensions, and (ii) \emph{inconsistent} prompts that assign conflicting values to the same feature. 



For each new prompt, we send the it back to GPT-5-mini for a self-refine~\citep{madaan2023self} to filter obvious errors and improve consistency. We then curate a subset of representative cases using three selection rules. Specifically, for each ambiguous prompt $\hat{p}$, we run ProCAD-coder to synthesize a CadQuery program and compute its CD against the ground-truth mesh. We keep a sample only if: (i) the original verified specification $p$ is high-quality, with $\mathrm{CD} < 2\times 10^{-4}$; (ii) the perturbed prompt $\hat{p}$ is genuinely harmful, with $\mathrm{CD} > 2\times 10^{-4}$; and (iii) the degradation is substantial: the ratio of the two Chamfer distances is at least 10. In addition, we include unambiguous prompts to balance the dataset. Using this pipeline, we construct a training set of 6{,}063 samples and a test set of 2{,}469 samples, with an approximately 1:1 ratio between unambiguous and ambiguous prompts. See Table~\ref{tab:ambig_stats_train_test} for detailed statistics.

We train ProCAD-clarifier using the same base model and fine-tuning hyperparameters as ProCAD-coder. With batch size 16, training takes 367 optimization steps and finishes in under 10 minutes. Table~\ref{tab:agentic_vs_single} reports the performance of our agentic system on the 2{,}469 test samples. We compare against two classes of baselines: (i) single-model, and (ii) agentic baselines . For all agentic variants, we keep the coding agent fixed as either our fine-tuned ProCAD-coder or Claude 4.5 Sonnet, since they achieve strong performance on unambiguous descriptions, as shown in Table~\ref{tab:testset_results_ir_cd}. 

Beyond Chamfer distance and invalidity ratio, we also evaluate the interaction quality using an LLM-as-judge for the two-agent system. Specifically, we compute an efficiency score that measures whether the clarifier's questions match the ground-truth questions without introducing redundant queries, and a resolution score that measures whether the clarified specification successfully resolves the ambiguity. See Appendix~\ref{app:mislead_metric} for more details. We include Cadrille~\citep{kolodiazhnyi2025cadrille} only for completeness, noting it was trained on standard Text2CAD data and lacks ambiguity detection capabilities. As this is the first work to explicitly address ambiguity in text-to-CAD, our primary baselines are general-purpose LLMs.
\vspace{-1mm}
\paragraph{Our two-agent system outperforms single-model systems}
Table~\ref{tab:agentic_vs_single} shows that introducing a clarification agent substantially improves robustness to ambiguous inputs. Compared to single-model direct coding, the two-agent pipeline consistently reduces the invalid rate and improves geometric fidelity, demonstrating the benefit of resolving underspecification and inconsistencies before code synthesis.
\vspace{-1mm}
\paragraph{ProCAD achieves the best overall results.}
Among all variants, pairing ProCAD-clarifier with ProCAD-coder yields the strongest performance across all metrics: it attains the lowest mean CD and invalid rate while also achieving the highest Efficiency and Resolution scores. Beyond geometric quality, ProCAD-clarifier minimizes user intervention by asking only the most necessary, targeted questions, and it produces higher-quality corrected prompts, which in turn enables more reliable downstream code generation. See Appendix~\ref{app:case_study_ambigous} for case studies where we fix the coding agent as {ProCAD-coder} and compare Claude Sonnet~4.5 and {ProCAD-clarifier} as the clarifying agent. See Appendix~\ref{app:images} for qualitative comparison.
\vspace{-1mm}
\paragraph{Generalization to out-of-distribution simulators.}
Moreover, we observe that ProCAD exhibits strong generalization capabilities, performing robustly even in out-of-distribution settings. While our training data consists entirely of user responses simulated by GPT-5-mini, we demonstrate in Table~\ref{tab:agentic_vs_single_haiku} that performance remains high when the user simulator is switched to Claude 4.5 Haiku~\citep{anthropic2025claude-haiku-4-5}. ProCAD consistently outperforms baselines where one of the agents is replaced by Claude 4.5 Sonnet.

\subsection{Human-Centered Evaluation}\label{section:human_study}

  The experiments in Tables~\ref{tab:agentic_vs_single} and~\ref{tab:agentic_vs_single_haiku} rely on LLM-based user
  simulators for both the ambiguous prompts and the user responses. To verify that the conclusions transfer to real
  users, we additionally conduct a human-centered evaluation in which both the ambiguous prompts and the user responses
  are provided by real annotators rather than LLM simulators.

  \paragraph{Setup.}
  We sample 100 examples from our high-quality Text-to-CadQuery dataset. For each example, we provide the original
  unambiguous prompt together with the reference image to human annotators, since the original prompt contains the
  precise dimensions. We then ask multiple annotators to rewrite ambiguous prompts. The resulting clarification
  questions and corrected prompts are further evaluated by CAD experts. In this human evaluation, we compare
  ProCAD-clarifier against three baselines while fixing the downstream generation model as ProCAD-coder for all methods.

  \paragraph{ProCAD-clarifier remains the strongest under real human interaction.}
  As shown in Table~\ref{tab:human_study}, ProCAD-clarifier continues to achieve the strongest performance across all
  metrics in this fully human-driven evaluation. Crucially, the interaction-quality metrics (Efficiency and Resolution)
  are now scored by human experts rather than an LLM judge, yet the conclusions are unchanged: our method consistently
  outperforms all three baselines. Since Resolution is closely tied to final geometric fidelity, the improvement in
  Chamfer Distance provides an additional proxy showing that the clarified prompts produced by ProCAD-clarifier better
  match the target shape.

  \paragraph{Reliability of LLM-based evaluation.}
  We also note that our LLM-based evaluation is intentionally designed to reduce judging difficulty. For Efficiency, the
   judge is asked to compare the clarification questions against the ground-truth missing information rather than make a
   fully open-ended quality judgment. For Resolution, we use a coarse three-level score $\{0, 0.5, 1\}$, which further
  reduces sensitivity. To directly quantify reliability, we randomly sample 100 test examples from the synthetic dataset
   and compare LLM-based evaluation against human evaluation of interaction quality. The two agree on more than $96\%$
  of the samples, providing additional evidence that the LLM-based metrics are reasonably reliable in our setting.

\section{Conclusion}
We introduced ProCAD, a novel two-agent framework that proactively addresses ambiguity in text-to-CAD generation. By fine-tuning our agents on a curated dataset of 10k high-quality samples, we demonstrated that resolving specification errors before code generation significantly improves reliability. Our text-to-CadQuery dataset suggests that description quality strongly affects downstream code synthesis: even with the same underlying geometry distribution, cleaner, more precise, and constraint-complete descriptions lead to markedly more reliable CAD programs. This highlights the need for expert-level, human-annotated datasets that reflect real engineering specifications in the future works. Our findings also underscore the necessity of moving beyond static prompting toward dynamic, interactive agents. Future directions include gathering large-scale ambiguity datasets from real-world human interactions and developing dedicated user-simulator models.








\section*{Acknowledgments}

We thank the anonymous reviewers for their valuable feedback and constructive suggestions, which helped improve the quality and clarity of this paper.

\section*{Impact Statement}
This paper presents work whose goal is to advance the field of Machine
Learning. There are many potential societal consequences of our work, none
which we feel must be specifically highlighted here.

\clearpage
\newpage

\bibliography{example_paper}

@article{briere2012comparing,
  title = {{Comparing 3D CAD models: uses, methods, tools and perspectives}},
  author={Bri{\`e}re-C{\^o}t{\'e}, Antoine and Rivest, Louis and Maranzana, Roland},
  journal={Computer-Aided Design and Applications},
  volume={9},
  number={6},
  pages={771--794},
  year={2012},
  publisher={Taylor \& Francis}
}

@InProceedings{Zhao_2024_CVPR,
    author    = {Zhao, Zelin and Fan, Fenglei and Liao, Wenlong and Yan, Junchi},
    title     = {Grounding and Enhancing Grid-based Models for Neural Fields},
    booktitle = {Proceedings of the IEEE/CVF Conference on Computer Vision and Pattern Recognition (CVPR)},
    month     = {June},
    year      = {2024},
    pages     = {19425-19435}
}

@inproceedings{zhao2021augmenting,
  title={Augmenting policy learning with routines discovered from a single demonstration},
  author={Zhao, Zelin and Gan, Chuang and Wu, Jiajun and Guo, Xiaoxiao and Tenenbaum, Joshua B},
  booktitle={Proceedings of the AAAI Conference on Artificial Intelligence},
  volume={35},
  number={12},
  pages={11024--11032},
  year={2021}
}

@inproceedings{NEURIPS2021_8d34201a,
 author = {Zhao, Zelin and Samel, Karan and Chen, Binghong and song, lee},
 booktitle = {Advances in Neural Information Processing Systems},
 editor = {M. Ranzato and A. Beygelzimer and Y. Dauphin and P.S. Liang and J. Wortman Vaughan},
 pages = {17021--17036},
 publisher = {Curran Associates, Inc.},
 title = {ProTo: Program-Guided Transformer for Program-Guided Tasks},
 url = {https://proceedings.neurips.cc/paper_files/paper/2021/file/8d34201a5b85900908db6cae92723617-Paper.pdf},
 volume = {34},
 year = {2021}
}

@article{robertson2002cad,
  title = {{CAD system use and engineering performance}},
  author={Robertson, David and Allen, Thomas J},
  journal={IEEE Transactions on Engineering Management},
  volume={40},
  number={3},
  pages={274--282},
  year={2002},
  publisher={IEEE}
}

@article{jiang2024survey,
  title = {{A survey on large language models for code generation}},
  author={Jiang, Juyong and Wang, Fan and Shen, Jiasi and Kim, Sungju and Kim, Sunghun},
  journal={arXiv preprint arXiv:2406.00515},
  year={2024}
}

@inproceedings{khan2024cad,
  title = {{Cad-signet: Cad language inference from point clouds using layer-wise sketch instance guided attention}},
  author={Khan, Mohammad Sadil and Dupont, Elona and Ali, Sk Aziz and Cherenkova, Kseniya and Kacem, Anis and Aouada, Djamila},
  booktitle = {{Proceedings of the IEEE/CVF Conference on Computer Vision and Pattern Recognition}},
  pages={4713--4722},
  year={2024}
}

@inproceedings{li2025cad,
  title = {{CAD-Llama: leveraging large language models for computer-aided design parametric 3D model generation}},
  author={Li, Jiahao and Ma, Weijian and Li, Xueyang and Lou, Yunzhong and Zhou, Guichun and Zhou, Xiangdong},
  booktitle = {{Proceedings of the Computer Vision and Pattern Recognition Conference}},
  pages={18563--18573},
  year={2025}
}

@article{chen2021evaluating,
  title = {{Evaluating large language models trained on code}},
  author={Chen, Mark},
  journal={arXiv preprint arXiv:2107.03374},
  year={2021}
}

@article{austin2021program,
  title = {{Program synthesis with large language models}},
  author={Austin, Jacob and Odena, Augustus and Nye, Maxwell and Bosma, Maarten and Michalewski, Henryk and Dohan, David and Jiang, Ellen and Cai, Carrie and Terry, Michael and Le, Quoc and others},
  journal={arXiv preprint arXiv:2108.07732},
  year={2021}
}

@article{badagabettu2024query2cad,
  title = {{Query2cad: Generating cad models using natural language queries}},
  author={Badagabettu, Akshay and Yarlagadda, Sai Sravan and Farimani, Amir Barati},
  journal={arXiv preprint arXiv:2406.00144},
  year={2024}
}

@inproceedings{li2024cad,
  title = {{Cad translator: An effective drive for text to 3d parametric computer-aided design generative modeling}},
  author={Li, Xueyang and Song, Yu and Lou, Yunzhong and Zhou, Xiangdong},
  booktitle = {{Proceedings of the 32nd ACM International Conference on Multimedia}},
  pages={8461--8470},
  year={2024}
}

@article{khan2024text2cad,
  title = {{Text2cad: Generating sequential cad designs from beginner-to-expert level text prompts}},
  author={Khan, Mohammad S and Sinha, Sankalp and Sheikh, Talha U and Stricker, Didier and Ali, Sk A and Afzal, Muhammad Z},
  journal={Advances in Neural Information Processing Systems},
  volume={37},
  pages={7552--7579},
  year={2024}
}

@article{xu2024cad,
  title = {{Cad-mllm: Unifying multimodality-conditioned cad generation with mllm}},
  author={Xu, Jingwei and Wang, Chenyu and Zhao, Zibo and Liu, Wen and Ma, Yi and Gao, Shenghua},
  journal={arXiv preprint arXiv:2411.04954},
  year={2024}
}

@inproceedings{wu2021deepcad,
  title = {{Deepcad: A deep generative network for computer-aided design models}},
  author={Wu, Rundi and Xiao, Chang and Zheng, Changxi},
  booktitle = {{Proceedings of the IEEE/CVF International Conference on Computer Vision}},
  pages={6772--6782},
  year={2021}
}

@article{guan2025cad,
  title = {{CAD-Coder: Text-to-CAD Generation with Chain-of-Thought and Geometric Reward}},
  author={Guan, Yandong and Wang, Xilin and Xing, Ximing and Zhang, Jing and Xu, Dong and Yu, Qian},
  journal={arXiv preprint arXiv:2505.19713},
  year={2025}
}

@article{jia2025meml,
  title = {{Meml-grpo: Heterogeneous multi-expert mutual learning for rlvr advancement}},
  author={Jia, Weitao and Lu, Jinghui and Yu, Haiyang and Wang, Siqi and Tang, Guozhi and Wang, An-Lan and Yin, Weijie and Yang, Dingkang and Nie, Yuxiang and Shan, Bin and others},
  journal={arXiv preprint arXiv:2508.09670},
  year={2025}
}

@article{kolodiazhnyi2025cadrille,
  title = {{cadrille: Multi-modal CAD Reconstruction with Online Reinforcement Learning}},
  author={Kolodiazhnyi, Maksim and Tarasov, Denis and Zhemchuzhnikov, Dmitrii and Nikulin, Alexander and Zisman, Ilya and Vorontsova, Anna and Konushin, Anton and Kurenkov, Vladislav and Rukhovich, Danila},
  journal={arXiv preprint arXiv:2505.22914},
  year={2025}
}

@article{xie2025text,
  title = {{Text-to-CadQuery: A New Paradigm for CAD Generation with Scalable Large Model Capabilities}},
  author={Xie, Haoyang and Ju, Feng},
  journal={arXiv preprint arXiv:2505.06507},
  year={2025}
}

@article{becattini2013introduction,
  title = {{About the introduction of a dialogue-based interaction within CAD systems}},
  author={Becattini, Niccol{\`o} and Borgianni, Yuri and Cascini, Gaetano and Rotini, Federico},
  journal={Computer-Aided Design and Applications},
  volume={10},
  number={3},
  pages={499--514},
  year={2013},
  publisher={Taylor \& Francis}
}

@article{zhang2024flexcad,
  title = {{Flexcad: Unified and versatile controllable cad generation with fine-tuned large language models}},
  author={Zhang, Zhanwei and Sun, Shizhao and Wang, Wenxiao and Cai, Deng and Bian, Jiang},
  journal={arXiv preprint arXiv:2411.05823},
  year={2024}
}

@article{alrashedy2024generating,
  title = {{Generating cad code with vision-language models for 3d designs}},
  author={Alrashedy, Kamel and Tambwekar, Pradyumna and Zaidi, Zulfiqar and Langwasser, Megan and Xu, Wei and Gombolay, Matthew},
  journal={arXiv preprint arXiv:2410.05340},
  year={2024}
}

@article{li2025seek,
  title = {{Seek-CAD: A Self-refined Generative Modeling for 3D Parametric CAD Using Local Inference via DeepSeek}},
  author={Li, Xueyang and Li, Jiahao and Song, Yu and Lou, Yunzhong and Zhou, Xiangdong},
  journal={arXiv preprint arXiv:2505.17702},
  year={2025}
}

@article{sun2025training,
  title = {{Training proactive and personalized llm agents}},
  author={Sun, Weiwei and Zhou, Xuhui and Du, Weihua and Wang, Xingyao and Welleck, Sean and Neubig, Graham and Sap, Maarten and Yang, Yiming},
  journal={arXiv preprint arXiv:2511.02208},
  year={2025}
}

@article{lu2024proactive,
  title = {{Proactive agent: Shifting llm agents from reactive responses to active assistance}},
  author={Lu, Yaxi and Yang, Shenzhi and Qian, Cheng and Chen, Guirong and Luo, Qinyu and Wu, Yesai and Wang, Huadong and Cong, Xin and Zhang, Zhong and Lin, Yankai and others},
  journal={arXiv preprint arXiv:2410.12361},
  year={2024}
}

@article{qing2025effibench,
  title = {{EffiBench-X: A Multi-Language Benchmark for Measuring Efficiency of LLM-Generated Code}},
  author={Qing, Yuhao and Zhu, Boyu and Du, Mingzhe and Guo, Zhijiang and Zhuo, Terry Yue and Zhang, Qianru and Zhang, Jie M and Cui, Heming and Yiu, Siu-Ming and Huang, Dong and others},
  journal={arXiv preprint arXiv:2505.13004},
  year={2025}
}

@article{niu2025intent,
  title = {{From Intent to Execution: Multimodal Chain-of-Thought Reinforcement Learning for Precise CAD Code Generation}},
  author={Niu, Ke and Yu, Haiyang and Chen, Zhuofan and Zhao, Mengyang and Fu, Teng and Li, Bin and Xue, Xiangyang},
  journal={arXiv preprint arXiv:2508.10118},
  year={2025}
}

@misc{cadquery_2_4_0_2024,
  author       = {{CadQuery Contributors}},
  title = {{CadQuery 2.4.0}},
  year         = {2024},
  howpublished = {\url{https://github.com/CadQuery/cadquery/releases/tag/2.4.0}},
  note         = {Accessed: 2026-01-25}
}

@inproceedings{cheong2014investigating,
  title = {{Investigating the use of controlled natural language as problem definition input for computer-aided design}},
  author={Cheong, Hyunmin and Li, Wei and Shu, LH and Bradner, Erin and Iorio, Francesco},
  booktitle = {{Proceedings of the 2014 International Conference on Innovative Design and Manufacturing (ICIDM)}},
  pages={65--70},
  year={2014},
  organization={IEEE}
}

@inproceedings{li2024llm4cad,
  title = {{LLM4CAD: Multi-Modal Large Language Models for 3D Computer-Aided Design Generation}},
  author={Li, Xingang and Sun, Yuewan and Sha, Zhenghui},
  booktitle = {{International Design Engineering Technical Conferences and Computers and Information in Engineering Conference}},
  volume={88407},
  pages={V006T06A015},
  year={2024},
  organization={American Society of Mechanical Engineers}
}

@article{govindarajan2025cadmium,
  title = {{Cadmium: Fine-tuning code language models for text-driven sequential cad design}},
  author={Govindarajan, Prashant and Baldelli, Davide and Pathak, Jay and Fournier, Quentin and Chandar, Sarath},
  journal={arXiv preprint arXiv:2507.09792},
  year={2025}
}

@article{an2026pr,
  title={PR-CAD: Progressive Refinement for Unified Controllable and Faithful Text-to-CAD Generation with Large Language Models},
  author={An, Jiyuan and Zhao, Jiachen and Chen, Fan and Yang, Liner and Liu, Zhenghao and Wang, Hongyan and An, Weihua and Zhang, Meishan and Yang, Erhong},
  journal={arXiv preprint arXiv:2604.19773},
  year={2026}
}

@inproceedings{rukhovich2025cad,
  title = {{Cad-recode: Reverse engineering cad code from point clouds}},
  author={Rukhovich, Danila and Dupont, Elona and Mallis, Dimitrios and Cherenkova, Kseniya and Kacem, Anis and Aouada, Djamila},
  booktitle = {{Proceedings of the IEEE/CVF International Conference on Computer Vision}},
  pages={9801--9811},
  year={2025}
}

@article{xu2022skexgen,
  title = {{Skexgen: Autoregressive generation of cad construction sequences with disentangled codebooks}},
  author={Xu, Xiang and Willis, Karl DD and Lambourne, Joseph G and Cheng, Chin-Yi and Jayaraman, Pradeep Kumar and Furukawa, Yasutaka},
  journal={arXiv preprint arXiv:2207.04632},
  year={2022}
}

@article{xu2023hierarchical,
  title = {{Hierarchical neural coding for controllable cad model generation}},
  author={Xu, Xiang and Jayaraman, Pradeep Kumar and Lambourne, Joseph G and Willis, Karl DD and Furukawa, Yasutaka},
  journal={arXiv preprint arXiv:2307.00149},
  year={2023}
}

@article{madaan2023self,
  title = {{Self-refine: Iterative refinement with self-feedback}},
  author={Madaan, Aman and Tandon, Niket and Gupta, Prakhar and Hallinan, Skyler and Gao, Luyu and Wiegreffe, Sarah and Alon, Uri and Dziri, Nouha and Prabhumoye, Shrimai and Yang, Yiming and others},
  journal={Advances in Neural Information Processing Systems},
  volume={36},
  pages={46534--46594},
  year={2023}
}

@inproceedings{wu2025inference,
  title = {{Inference scaling laws: An empirical analysis of compute-optimal inference for LLM problem-solving}},
  author={Wu, Yangzhen and Sun, Zhiqing and Li, Shanda and Welleck, Sean and Yang, Yiming},
  booktitle = {{The Thirteenth International Conference on Learning Representations}},
  year={2025}
}

@inproceedings{snell2025scaling,
  title = {{Scaling LLM test-time compute optimally can be more effective than scaling parameters for reasoning}},
  author={Snell, Charlie Victor and Lee, Jaehoon and Xu, Kelvin and Kumar, Aviral},
  booktitle = {{The Thirteenth International Conference on Learning Representations}},
  year={2025}
}

@inproceedings{shi2025judging,
  title = {{Judging the judges: A systematic study of position bias in llm-as-a-judge}},
  author={Shi, Lin and Ma, Chiyu and Liang, Wenhua and Diao, Xingjian and Ma, Weicheng and Vosoughi, Soroush},
  booktitle = {{Proceedings of the 14th International Joint Conference on Natural Language Processing and the 4th Conference of the Asia-Pacific Chapter of the Association for Computational Linguistics}},
  pages={292--314},
  year={2025}
}

@article{zheng2023judging,
  title = {{Judging llm-as-a-judge with mt-bench and chatbot arena}},
  author={Zheng, Lianmin and Chiang, Wei-Lin and Sheng, Ying and Zhuang, Siyuan and Wu, Zhanghao and Zhuang, Yonghao and Lin, Zi and Li, Zhuohan and Li, Dacheng and Xing, Eric and others},
  journal={Advances in neural information processing systems},
  volume={36},
  pages={46595--46623},
  year={2023}
}

@article{saito2023verbosity,
  title = {{Verbosity bias in preference labeling by large language models}},
  author={Saito, Keita and Wachi, Akifumi and Wataoka, Koki and Akimoto, Youhei},
  journal={arXiv preprint arXiv:2310.10076},
  year={2023}
}

@article{dubois2024length,
  title = {{Length-controlled alpacaeval: A simple way to debias automatic evaluators}},
  author={Dubois, Yann and Galambosi, Bal{\'a}zs and Liang, Percy and Hashimoto, Tatsunori B},
  journal={arXiv preprint arXiv:2404.04475},
  year={2024}
}

@misc{anthropic2025claude-sonnet-4-5,
  author       = {{Anthropic}},
  title = {{Introducing Claude Sonnet 4.5}},
  year         = {2025},
  month        = {September},
  howpublished = {\url{https://www.anthropic.com/news/claude-sonnet-4-5}},
  note         = {Accessed: 2026-01-25}
}

@article{sekulic2024reliable,
  title = {{Reliable LLM-based user simulator for task-oriented dialogue systems}},
  author={Sekuli{\'c}, Ivan and Terragni, Silvia and Guimar{\~a}es, Victor and Khau, Nghia and Guedes, Bruna and Filipavicius, Modestas and Manso, Andre Ferreira and Mathis, Roland},
  journal={arXiv preprint arXiv:2402.13374},
  year={2024}
}

@inproceedings{zhang2025llm,
  title = {{Llm-powered user simulator for recommender system}},
  author={Zhang, Zijian and Liu, Shuchang and Liu, Ziru and Zhong, Rui and Cai, Qingpeng and Zhao, Xiangyu and Zhang, Chunxu and Liu, Qidong and Jiang, Peng},
  booktitle = {{Proceedings of the AAAI Conference on Artificial Intelligence}},
  volume={39},
  number={12},
  pages={13339--13347},
  year={2025}
}

@inproceedings{ito2025enhancing,
  title = {{Enhancing Proactive Dialogue Systems Through Self-Learning of Reasoning and Action-Planning}},
  author={Ito, Ryosuke and Takiguchi, Tetsuya and Ariki, Yasuo},
  booktitle = {{Proceedings of the 15th International Workshop on Spoken Dialogue Systems Technology}},
  pages={165--171},
  year={2025}
}

@inproceedings{sahay2025ask,
  title = {{Ask: Aspects and retrieval based hybrid clarification in task oriented dialogue systems}},
  author={Sahay, Rishav and Tekumalla, Lavanya Sita and Aggarwal, Purav and Jain, Arihant and Saladi, Anoop},
  booktitle = {{Proceedings of the 63rd Annual Meeting of the Association for Computational Linguistics (Volume 6: Industry Track)}},
  pages={881--895},
  year={2025}
}

@article{testoni2024asking,
  title = {{Asking the right question at the right time: Human and model uncertainty guidance to ask clarification questions}},
  author={Testoni, Alberto and Fern{\'a}ndez, Raquel},
  journal={arXiv preprint arXiv:2402.06509},
  year={2024}
}

@inproceedings{ngiam2011multimodal,
  title = {{Multimodal deep learning.}},
  author={Ngiam, Jiquan and Khosla, Aditya and Kim, Mingyu and Nam, Juhan and Lee, Honglak and Ng, Andrew Y and others},
  booktitle = {{ICML}},
  volume={11},
  pages={689--696},
  year={2011}
}

@article{singh2025openai,
  title = {{Openai gpt-5 system card}},
  author={Singh, Aaditya and Fry, Adam and Perelman, Adam and Tart, Adam and Ganesh, Adi and El-Kishky, Ahmed and McLaughlin, Aidan and Low, Aiden and Ostrow, AJ and Ananthram, Akhila and others},
  journal={arXiv preprint arXiv:2601.03267},
  year={2025}
}

@article{bai2025qwen2,
  title = {{Qwen2. 5-vl technical report}},
  author={Bai, Shuai and Chen, Keqin and Liu, Xuejing and Wang, Jialin and Ge, Wenbin and Song, Sibo and Dang, Kai and Wang, Peng and Wang, Shijie and Tang, Jun and others},
  journal={arXiv preprint arXiv:2502.13923},
  year={2025}
}

@article{liu2024deepseek,
  title = {{Deepseek-v3 technical report}},
  author={Liu, Aixin and Feng, Bei and Xue, Bing and Wang, Bingxuan and Wu, Bochao and Lu, Chengda and Zhao, Chenggang and Deng, Chengqi and Zhang, Chenyu and Ruan, Chong and others},
  journal={arXiv preprint arXiv:2412.19437},
  year={2024}
}

@misc{anthropic2025claude-haiku-4-5,
  author       = {{Anthropic}},
  title = {{Introducing Claude Haiku 4.5}},
  year         = {2025},
  month        = {October},
  howpublished = {\url{https://www.anthropic.com/news/claude-haiku-4-5}},
  note         = {Accessed: 2026-01-25}
}

@article{wang2025text,
  title={Text-to-cad generation through infusing visual feedback in large language models},
  author={Wang, Ruiyu and Yuan, Yu and Sun, Shizhao and Bian, Jiang},
  journal={arXiv preprint arXiv:2501.19054},
  year={2025}
}
\bibliographystyle{icml2026}

\newpage
\appendix
\onecolumn

\paragraph {Appendix Summary} This appendix provides supplementary details and qualitative analysis to support the main findings. We first show the extended related works in Appendix~\ref{app:related_work}. We then analyze specific failure modes associated with explicit scaling operations in Appendix~\ref{app:failure_mode} and present a qualitative case study comparing our natural language descriptions against the Text2CAD baseline in Appendix~\ref{app:case_study}. We then report detailed statistics on the ground-truth CadQuery code quality (Appendix~\ref{app:statitics_Chamfer_distance}) and the distribution of the ambiguous prompt dataset (Appendix~\ref{app:mislead_dataset_statis}). Furthermore, Appendix~\ref{app:train_set_up} compares our experimental setup  with related works, while Appendix~\ref{app:mislead_metric} defines the exact LLM-as-judge metrics used to evaluate the clarification agent. Appendix~\ref{app:case_study_ambigous} lists examples where Our ProCAD outperforms Claude Sonnet 4.5 on resolving. Appendix~\ref{app:images} shows qualitative comparison against baselines. Then Appendix~\ref{app:failure_mode_clarifier} shows the failure mode of our system. Appendix~\ref{app:protocol_ablation} shows that our protocol balances clarification quality and latency best. Finally, Appendix~\ref{app:system_prompts} provides the full set of system prompts used for inference, data annotation, and ambiguity generation.

\section{Related Works}\label{app:related_work}

\subsection{Text-to-CAD and Parametric Shape Generation}
Early learning-based CAD generation methods focused on synthesizing shapes from structured representations such as voxel grids, meshes, or boundary representations, often without explicit programmatic edit-ability~\citep{wu2021deepcad,Zhao_2024_CVPR}. More recent work has emphasized generating parametric CAD models represented as command sequences or sketch-extrude programs, enabling downstream modification and reuse~\citep{khan2024cad}. These approaches typically assume access to clean, fully specified input signals, such as aligned sketches or reference shapes~\citep{khan2024cad}.

With the rise of natural-language interfaces, text-to-CAD has emerged as a promising direction for lowering the barrier to CAD modeling~\citep{badagabettu2024query2cad,li2024cad,khan2024text2cad}. Existing methods generally cast text-to-CAD as a conditional generation problem, mapping user prompts directly to CAD programs or intermediate representations via supervised learning. While these methods demonstrate impressive results under curated benchmarks, they often struggle when prompts are ambiguous, incomplete, or internally inconsistent—a common case in real-world human descriptions~\citep{becattini2013introduction}.

\subsection{LLMs for CAD Generation}
Recent advances in large language models (LLMs) for program synthesis~\citep{zhao2021augmenting,NEURIPS2021_8d34201a,chen2021evaluating,austin2021program,jiang2024survey} have motivated their application to CAD code generation. Several works leverage LLMs to translate natural language into structured CAD scripts, including OpenSCAD, CadQuery, or proprietary CAD-like languages~\citep{xie2025text,guan2025cad,kolodiazhnyi2025cadrille,jia2025meml}. Among these, CadQuery has gained particular traction due to its Python-based syntax and compositional structure, which aligns well with LLMs’ strong performance on Python code generation~\citep{cadquery_2_4_0_2024,niu2025intent,qing2025effibench}.

Most existing text-to-CadQuery systems follow a one-shot generation paradigm, either using prompt-engineered frontier models or fine-tuned open-source LLMs~\citep{kolodiazhnyi2025cadrille,guan2025cad}. Some works incorporate execution feedback or geometric validation to iteratively refine generated code~\citep{alrashedy2024generating,li2025seek}. However, these approaches still treat the original user prompt as a fixed specification and rely on post-hoc correction when failures occur. In contrast, our work addresses specification errors \emph{before} code generation by explicitly auditing and completing the textual description, thereby reducing downstream failure modes.

\subsection{Text-to-CadQuery Dataset}\label{section:related_work_dataset}

To the best of our knowledge, only a small number of prior works provide \emph{both} expert-level natural-language descriptions and ground-truth CadQuery programs. While some datasets include general text descriptions without precise dimensions~\citep{xu2024cad,khan2024text2cad}, producing executable CadQuery code typically requires expert-level, highly precise specifications, making large-scale annotation costly and difficult to scale.

 LLM4CAD~\citep{li2024llm4cad} contains roughly 5,000 annotated samples, but it focuses on only five common mechanical part categories. Query4CAD~\citep{badagabettu2024query2cad} is substantially smaller, with just 57 samples. Consequently, many text-to-CadQuery studies build their own annotation pipelines by prompting an LLM or VLM and filtering low-quality programs\citep{xie2025text,guan2025cad}, primarily using the expert-level procedural instructions in Text2CAD~\citep{khan2024text2cad} as the primary text source. However, translating Text2CAD's expert descriptions into CadQuery code is nontrivial: these descriptions are frequently noisy and overly long, containing redundant details that can distract the model and increase the risk of hallucinated or incorrect code~\citep{govindarajan2025cadmium}. In particular, scaling operations frequently result in misleading descriptions, as documented in the failure mode analysis in Appendix~\ref{app:failure_mode}. Although \citet{kolodiazhnyi2025cadrille} pair expert descriptions directly with the CadQuery code from \citet{rukhovich2025cad}, their approach overlooks the discrepancy between the units and commands specified in the text versus the actual code: in Text2CAD, key dimensions and units are derived from the minimal JSON specification, whereas the CadQuery programs are reconstructed from command sequences. This discrepancy makes it difficult for LLMs to align text and code reliably, especially in a zero-shot setting.



\subsection{Proactive and Agentic Language Models}
Proactive agents extend reactive instruction-following models by anticipating user needs, identifying missing or inconsistent information, and initiating clarification before acting~\citep{lu2024proactive,sun2025training}. Such agentic behaviors have shown benefits in task-oriented dialogue, decision support, and program synthesis, where agents may decompose tasks, validate intermediate results, or iteratively refine specifications. However, their application to geometric modeling and CAD remains limited. We bring proactive agent design into text-to-CAD generation by introducing a dedicated specification agent that audits prompts for completeness and consistency and interacts with the user only when necessary. Trained with domain-specific supervision derived from systematically perturbed CAD specifications, our agent balances robustness to ambiguous prompts with low interaction overhead.

\section{Failure modes of scaling operations in Text2CAD}~\label{app:failure_mode}
\begin{figure}[!ht]
\centering
\begin{tcolorbox}[
  width=0.95\linewidth,
  colback=white,
  colframe=black,
  boxrule=0.6pt,
  arc=1.5mm,
  left=6pt,right=6pt,top=6pt,bottom=6pt
]
\textbf{Original prompt (excerpt):}
\begin{quote}\itshape
... create a circle with center \textbf{(0.1293, 0.1293)} and radius \textbf{0.1293}. Then add another concentric circle with radius \textbf{0.0853}. After completing the sketch, \textbf{apply a scaling factor of 0.2586 to the entire sketch}. ... \textbf{extrude} the sketch by \textbf{0.75} units along the normal direction. The dimensions of the resulting object are \textbf{0.2586} in length, \textbf{0.2586} in width, and \textbf{0.75} in height.
\end{quote}

\vspace{2mm}
\hrule
\vspace{2mm}

\textbf{Incorrect CadQuery code:}
\vspace{1mm}

\begin{lstlisting}[language=Python,basicstyle=\ttfamily\footnotesize]
import cadquery as cq

wp = (cq.Workplane("XY")
      .center(0.1293, 0.1293)
      .circle(0.1293)
      .circle(0.0853))

wp = wp.scale(0.2586)  # incorrect: Workplane has no scale() API
result = wp.extrude(0.75)
\end{lstlisting}
\end{tcolorbox}
\vspace{-1mm}
\caption{One failure example in Text2CAD for text-to-CadQuery generation.}
\label{fig:text2cad-scale-prompt}
\end{figure}

We observe that explicit scaling operations appear in a large fraction of Text2CAD examples. This design choice is historically motivated by earlier sequence-based CAD generation settings, where Transformer models were assumed to represent continuous parameters within a fixed numeric range; consequently, real-valued dimensions were rescaled into a predefined interval to ease tokenization and command-sequence prediction. However, this convention transfers poorly to CadQuery, where geometry is expressed as executable Python code. In CadQuery, scaling is not a generic operation that can be freely applied at any stage: in particular, a naive interpretation such as ``scale the workplane'' may prompt weaker code generators to hallucinate unsupported APIs, e.g.,\texttt{Workplane.scale(...)}, yielding invalid programs. We further note that the \emph{ordering} in these prompts makes mistakes particularly likely: the scaling step is described immediately after the 2D sketch construction, rather than after the 3D solid is created via extrusion . This placement encourages an implementation that attempts to scale the workplane. Therefore, the prompt structure itself can systematically bias code generators toward an invalid code, even when the underlying geometry is simple.

Moreover, as illustrated in Figure~\ref{fig:text2cad-scale-prompt}, the scaling statement admits two plausible but conflicting interpretations:
\begin{itemize}
  \item \textbf{Interpretation A (literal post-sketch scaling).}
  One may follow the prompt literally by first drawing the circles with center $(0.1293, 0.1293)$ and radii $0.1293$ and $0.0853$, and then scaling the entire sketch by $0.2586$. Under this interpretation, the outer radius becomes $0.2586 \cdot 0.1293 \approx 0.0334$, implying a footprint of approximately $2 \times 0.2586 \cdot 0.1293 \approx 0.0669$. This contradicts the stated final dimensions $0.2586 \times 0.2586 \times 0.75$.

  \item \textbf{Interpretation B (parameters already in the target units).}
  Alternatively, one may treat the listed coordinates and radii as already expressed in the final unit system, in which case the explicit scaling step is redundant and should be ignored.
\end{itemize}


\section{Case Study: Comparing Text2CAD with our descriptions}\label{app:case_study}

\begin{tcolorbox}[
  title=\textbf{Text Descriptions for a Simple Rectangular Prism},
  colback=gray!10!white,
  colframe=black,
  breakable
]
\small

\textbf{Text2CAD}

Create a new coordinate system by setting the Euler angles to negative ninety degrees, zero degrees, and negative ninety degrees, and set the translation vector to zero, zero, and zero. On the first face, draw the first loop consting of four straight lines. The first line starts at the origin (zero, zero) and ends at (zero point 0.75, zero). The second line starts at (zero 0.75, zero) and ends at (zero point, zero 0.5566225). The third line starts at (zero 0.75, 0.5566225) and ends at (zero, zero 0.556625). The fourth line starts at (zero, zero, 0.556225) and returns to the origin. Scale the entire sketch by a factor of zero point .75. Rotate the scaled sketch using the same Euler angles as the coordinate system and translate it using the same translation vector. Extrude the transformed sketch sketch by zero point 0.2281 units along the normal direction to create a new solid body. The final dimensions of the rectangular prism with rounded edges and a slightly tilted top surface are: length zero 0.75 units, width zero 0.556225 units, and height zero 0.02801 125 units.

\vspace{1mm}
\noindent\rule{\linewidth}{0.4pt}
\vspace{1mm}

\textbf{Ours}

This is a single solid rectangular plate: a plain 200 by 150 face extruded to a thickness of 7 (no holes or additional features).

Setup: workplane is the YZ plane with its origin translated to (-4, -100, -75).

Build: on that YZ workplane sketch a rectangle 200 (along the workplane's first axis) by 150 (along the workplane's second axis) with its lower-left corner at the workplane origin (sketch points at (0,0) to (200,150)). Extrude the rectangle 7 in the positive normal direction to form the solid plate.

\vspace{1mm}
\noindent\rule{\linewidth}{0.4pt}
\vspace{1mm}

\textbf{CadQuery Code }
\begin{verbatim}
import cadquery as cq

w0 = cq.Workplane('YZ', origin=(-4, -100, -75))

r = w0.sketch().face(
    w0.sketch()
      .segment((0,0), (200,0))
      .segment((200,0), (200,150))
      .segment((200,150), (0,150))
      .segment((0,150), (0,0))
      .assemble()
).finalize().extrude(7)
\end{verbatim}

\end{tcolorbox}

We notice that even for a simple rectangular prism, the Text2CAD description is overly long and unnatural, whereas our generated prompt is more concise and precisely captures the shape specification. Among the three representations (expert-level text in Text2CAD, CadQuery code, and our text), Text2CAD is verbose, CadQuery is concise but abstract, and our representation is both human-like and compact while remaining easy to interpret.

\section{Data Statistics of CadQuery code}\label{app:statitics_Chamfer_distance}

We report Chamfer Distance statistics for the CadQuery programs released by~\citet{rukhovich2025cad}. As shown in Table~\ref{tab:cd_stats}, the vast majority of samples reconstruct the target geometry with high fidelity, indicating that these programs can serve as reliable ground-truth code. This provides a practical alternative to relying on the minimal structured JSON representation used in Text2CAD. In particular, over $93\%$ of samples achieve a Chamfer Distance below $2\times 10^{-4}$. 

\begin{table}[!ht]
\centering
\caption{Chamfer distance distribution for CadQuery reconstructions from~\citet{rukhovich2025cad}. Percentage denotes the fraction of samples whose Chamfer distance is below the specified threshold.}
\label{tab:cd_stats}
\begin{tabular}{l r}
\hline
\textbf{CD ($\times 10^{3}$)} & \textbf{Percentage} \\
\hline
$0.1$   & 69.81\% \\
$0.2$   & 93.37\% \\
$0.5$   & 97.95\% \\
$1$     & 98.63\% \\
$2$     & 99.07\% \\
\hline
\end{tabular}
\end{table}

\section{Data Statistics of Ambiguous Prompts}~\label{app:mislead_dataset_statis}

\begin{table}[!ht]
\centering
\caption{Train and test split statistics for different ambiguity types. Here, the number of issues refers to the number of dimensions that contain ambiguities.}
\label{tab:ambig_stats_train_test}
\begin{tabular}{lcc}
\toprule
 & \textbf{Train (N=6,063)} & \textbf{Test (N=2,469)} \\
\midrule
Unambiguous  & 3,200 & 1,000 \\
\midrule
Under-specified (1 issue) & 1,071 &   427 \\
Under-specified (2 issues) &   479 &   638 \\
\textbf{total}             & \textbf{1,550} & \textbf{1,065} \\
\midrule
Conflicting (1 issue)     &   989 &   314 \\
Conflicting (2 issues)     &   324 &    90 \\
\textbf{total}             & \textbf{1,313} & \textbf{404} \\
\bottomrule
\end{tabular}
\end{table}

\section{Setup Comparison with existing Text-to-CadQuery works}\label{app:train_set_up}

\begin{table}[!ht]
\centering
\small
\caption{Training setup and data sources for Text-to-CadQuery.}
\label{tab:cadcoder-setup}
\setlength{\tabcolsep}{1pt}
\renewcommand{\arraystretch}{1.2}

\begin{tabular}{l c c c c c}
\hline
Model & Train \# & Base model & Training & Text source & CadQuery source \\
\hline
ProCAD-coder (ours) & 1.6K & Qwen2.5-7B-Instruct & SFT & GPT-5-mini & CAD-recode \\
Cadrille~\citep{kolodiazhnyi2025cadrille} & 160K & Qwen2-VL-2B & SFT+RL & Text2CAD & CAD-recode \\
PR-CAD~\citep{an2026pr} & 150K & Qwen2.5-7B-Instruct & SFT+RL & Qwen2.5-72B & Gemini-2.5-Flash \\
Text2CadQuery~\citep{xie2025text} & 150K & Qwen2.5-3B & SFT & Text2CAD & Gemini-2.0-Flash \\
CAD-coder~\citep{guan2025cad} & 150K & Qwen2.5-7B-Instruct & SFT+RL & Text2CAD & DeepSeek-V3~\citep{liu2024deepseek} \\
\hline
\end{tabular}
\vspace{1mm}
\end{table}

\begin{table}[!ht]
\centering
\caption{Performance comparison. CD values are scaled by $10^3$; lower is better.}
\label{tab:cadcoder-performance}
\footnotesize
\setlength{\tabcolsep}{6pt}
\renewcommand{\arraystretch}{1.2}
\begin{tabular}{l c c c}
\hline
Model & Mean CD ($\times 10^3$)$\downarrow$ & Median CD ($\times 10^3$)$\downarrow$ & IR (\%)$\downarrow$ \\
\hline
ProCAD-coder (ours) & 0.108 & 0.066 & 0.9 \\
Cadrille~\citep{kolodiazhnyi2025cadrille} & -- & 0.17 & 0.0 \\
PR-CAD~\citep{an2026pr} & 5.87 & -- & 0.62 \\
Text2CadQuery~\citep{xie2025text} & 10.229 & 0.191 & 6.5 \\
CAD-coder~\citep{guan2025cad} & 6.54 & 0.17 & 1.45 \\
\hline
\end{tabular}
\vspace{1mm}

\end{table}

Here, we also summarize the training setup and reported performance of representative text-to-CadQuery systems. In Table~\ref{tab:cadcoder-setup}, {Text Source} indicates where the textual descriptions come from, and {CadQuery Source} indicates where the CadQuery programs come from. Note that the number of training samples for {PR-CAD} is taken from its rebuttal on OpenReview rather than the main paper. Notably, while several prior works rely on more than 150K training samples, our approach achieves strong results with only {1.6K} samples using standard SFT. We also include the performance numbers reported in the original papers in Table~\ref{tab:cadcoder-performance}; however, because the test sets differ across works, these results are not directly comparable and are provided only for completeness.

\section{LLM-as-judge Metrics for the clarifying
agent}\label{app:mislead_metric}

We design two metrics to evaluate the communication quality and ambiguity resolution ability of the clarifying agent with an efficiency score and a resolution score, respectively. For unambiguous prompts, if the clarification agent incorrectly flags the prompt as ambiguous, we assign both scores to $0$; if it correctly marks the prompt as unambiguous, we assign both scores to $1$. Similarly, for ambiguous prompts, if the agent incorrectly marks the prompt as unambiguous, we assign both scores to $0$. For all other cases, we use the following LLM-based judge to measure the scores.

\paragraph{Efficiency}
We cast evaluation as a set matching problem and use an LLM-as-judge to align generated questions to ground-truth questions:
\begin{itemize}
  \item A generated question is counted as a \textbf{match} if  there is a semantically equavalent ground-truth question.
  \item Any generated question that does not match any ground-truth question is marked as \textbf{redundant}.
\end{itemize}
Based on the matching, we compute standard precision and recall over questions and define the efficiency score as the F1 measure.
A higher efficiency indicates that the agent asks the right questions while avoiding redundant questions.

\paragraph{Precision (resolution quality).}
Let $p^{\star}$ be the target unambiguous specification (ground truth) and let $\hat{p}$ be the clarified specification produced by the agent after incorporating the user's answers. We ask an LLM-as-judge to compare $\hat{p}$ against $p^{\star}$ and output a discrete resolution score:
\[
\text{Precision}(\hat{p},p^{\star}) \in \{0,\; 0.5,\; 1\},
\]
where:
\begin{itemize}
  \item $1$: the ambiguity is fully resolved and the clarified prompt is consistent with the target specification;
  \item $0.5$: partially resolved, e.g., when a sample contains multiple issues and only a subset is correctly fixed;
  \item $0$: not resolved.
\end{itemize}

Here are the system prompts for both metrics.

\begin{tcolorbox}[title=\textbf{System Prompt: Efficiency Judging}, colback=gray!10!white, colframe=black, breakable]
\small
You are an impartial logic evaluator. Determine whether a set of \textbf{Generated Questions} maps correctly to the \textbf{Ground Truth Questions}. You must categorize every generated question into one of two lists:
\begin{itemize}\setlength\itemsep{0.15em}
  \item \textbf{Matched:} the question asks for the same variable/dimension as a ground-truth question.
  \item \textbf{Hallucinated:} the question asks for something irrelevant, incorrect, or not present in the ground truth.
\end{itemize}

\vspace{0.4em}
\textbf{Criteria for a match.}
\begin{itemize}\setlength\itemsep{0.15em}
  \item The intent must be identical (asking for the same  geometric feature).
  \item Phrasing differences are allowed.
\end{itemize}

\vspace{0.6em}
\textbf{Output format (strictly valid JSON).}
\begin{small}
\begin{verbatim}
{
  "hallucinated_questions": [
    "<list of generated questions that do NOT match any ground truth>"
  ],
  "matched_questions": [
    {
      "generated_question": "<the generated question>",
      "matched_ground_truth": "<the specific ground truth question it corresponds to>"
    }
  ]
}
\end{verbatim}
\end{small}

\end{tcolorbox}

\begin{tcolorbox}[title=\textbf{System Prompt: Resolution Judging}, colback=gray!10!white, colframe=black, breakable]
\small
You are a CAD specification auditor. You must compare a ``CLARIFIED PROMPT'' against the ``ORIGINAL GROUND TRUTH'' to see if ambiguities have been resolved correctly. You must assign a \texttt{resolution\_status} score based strictly on these rules:

\textbf{SCORE 1.0 (Fully Resolved):}
\begin{itemize}\setlength\itemsep{0.15em}
  \item All missing dimensions, coordinates, or specifications from the Original have been restored.
  \item The values match the Original Ground Truth exactly (or are mathematically equivalent).
  \item No conflicting information remains.
\end{itemize}

\textbf{SCORE 0.5 (Partially Resolved):}
\begin{itemize}\setlength\itemsep{0.15em}
  \item Some missing info was restored, but other parts are still missing or incorrect.
  \item OR the model fixed the ambiguity but used the wrong values (e.g., guessed a dimension instead of getting the correct one).
\end{itemize}

\textbf{SCORE 0.0 (Unresolved/Failed):}
\begin{itemize}\setlength\itemsep{0.15em}
  \item The prompt remains ambiguous.
  \item OR the model introduced new conflicts.
  \item OR the model hallucinated critical details that make the part invalid.
\end{itemize}

\textbf{Output JSON:}
\begin{verbatim}
{
  "score": <0.0, 0.5, or 1.0>,
  "reasoning": "<concise explanation of why it fits this category>"
}
\end{verbatim}
\end{tcolorbox}

\section{Case study on resolving ambiguous prompts}\label{app:case_study_ambigous}

In these examples, using Claude Sonnet 4.5 as the clarification agent either fails to produce a corrected specification or asks redundant questions, whereas ProCAD identifies the key ambiguity and generates the corrected prompt; in all cases, the coding agent is fixed as ProCAD-coder.

\begin{tcolorbox}[
  title=\textbf{Case Study: Redundant Clarification},
  colback=gray!10!white,
  colframe=black,
  breakable
]
\small

\textbf{Ambiguous prompt}

This is a rectangular mounting plate 200 $\times$ 160 with a uniform thickness of 20 and four circular through-holes (radius 8) positioned on the face.

Workplane: sketch on the XY plane with the workplane origin moved to $(-100,-80,-10)$.

Build: on that XY workplane sketch a rectangle with corners at $(0,0)$ and $(200,160)$. Add circular cutouts \textbf{(radius 10)} centered at $(40,20)$, $(40,120)$, $(90,80)$, and $(120,20)$. Extrude the sketch 20 to form the plate.

\vspace{1mm}
\noindent\rule{\linewidth}{0.4pt}
\vspace{1mm}

\textbf{Ground-Truth Ambiguity}

There is a {direct conflict} in the specification of a single geometric feature. The through-hole radius is specified as 8 in the description. The build instructions specify a radius of 10.

\vspace{1mm}
\noindent\rule{\linewidth}{0.4pt}
\vspace{1mm}

\textbf{Claude 4.5 Sonnet Questions}

\begin{itemize}
    \item "There is a conflict in the hole radius specification. The description states 'radius 8' but the build instructions specify 'radius 10' for the circular cutouts. Which radius should be used for the through-holes?",
    \item "There is a conflict in the hole positions. The description mentions 'four circular through-holes' but only three unique positions are clearly specified in the build instructions: (40,20), (40,120), (90,80), and (120,20) - however, this is actually four positions. Can you confirm these four hole positions are correct: (40,20), (40,120), (90,80), and (120,20)?"
\end{itemize}
\vspace{1mm}
\noindent\rule{\linewidth}{0.4pt}
\vspace{1mm}

\textbf{ProCAD-clarifier Questions}

\begin{itemize}
    \item "For the circular through-holes, should the radius be 8 or 10?"

\end{itemize}
\end{tcolorbox}
As shown in this example, Claude 4.5 Sonnet correctly identifies the radius conflict but additionally asks the user to reconfirm hole positions that are already fully specified, resulting in redundant interaction without resolving new ambiguity. In contrast, ProCAD-clarifier asks a single, targeted question that directly addresses the only conflicting feature, resolving the ambiguity with minimal user effort.

\begin{tcolorbox}[
  title=\textbf{Case Study: Missed Necessary Clarification},
  colback=gray!10!white,
  colframe=black,
  breakable
]
\small

\textbf{Ambiguous Prompt}

This is a solid cylindrical rod (a single extruded circle) with a circular cross-section and a length of 200. The cylinder's axis is aligned with the workplane normal (i.e., normal to the XY plane).

Setup: work on the XY workplane whose origin has been shifted.

Build description: On that shifted XY workplane, sketch a circle whose center is placed at $(19,0)$ relative to the workplane origin. Finalize the sketch and extrude the circle 200 in the positive normal direction to produce the cylinder.

\vspace{1mm}
\noindent\rule{\linewidth}{0.4pt}
\vspace{1mm}

\textbf{Ground-Truth Ambiguity}

The prompt is {under-specified}. The coordinates of the shifted workplane origin are missing. Also  the radius of the sketched circle is not specified.

\vspace{1mm}
\noindent\rule{\linewidth}{0.4pt}
\vspace{1mm}

\textbf{Claude 4.5 Sonnet}

Claude 4.5 Sonnet incorrectly classifies the prompt as unambiguous and asks no clarification questions, proceeding directly to code generation. This results in a large geometric error due to unresolved missing parameters.

\vspace{1mm}
\noindent\rule{\linewidth}{0.4pt}
\vspace{1mm}

\textbf{ProCAD-clarifier Questions}

\begin{itemize}
    \item ``What are the coordinates of the shifted workplane origin (the exact shift vector)?''
    \item ``What radius should the sketched circle have?''
\end{itemize}

\textbf{User Answers}
1. Shifted workplane origin: (-19, 0, -100).  2. Circle radius: 19.

\textbf{ProCAD-clarifier Corrected Prompt }

This is a solid cylindrical rod (a single extruded circle) with a circular cross-section of radius 19 and a length of 200. The cylinder's axis is aligned with the workplane normal.

Setup: work on the XY workplane whose origin has been shifted to $(-19, 0, -100)$.

Build description: On that shifted XY workplane, sketch a circle of radius 19 whose center is placed at $(19,0)$ relative to the workplane origin. Finalize the sketch and extrude the circle 200 in the positive normal direction to produce the cylinder.

\end{tcolorbox}

This example illustrates the opposite failure mode of redundant clarification: Claude 4.5 Sonnet fails to detect genuine under-specification and skips necessary clarification entirely, leading to incorrect geometry. In contrast, ProCAD-clarifier precisely identifies the missing parameters, asks only the required questions, and fully recovers the correct specification before code generation. In this case, Claude 4.5 fails to detect the ambiguity and passes the prompt directly to ProCAD-coder, the resulting Chamfer distance is $7.56\times10^{-3}$; in contrast, our ProCAD-clarifier resolves the missing details first, and with the same ProCAD-coder as the coding agent achieves $2.68\times10^{-5}$.

\begin{tcolorbox}[
  title=\textbf{Case Study: Missed Clarification for an L-shaped Extrusion},
  colback=gray!10!white,
  colframe=black,
  breakable
]
\small

\textbf{Ambiguous Prompt}

This is a prismatic L-shaped block (stepped L profile) formed by extruding a closed polygon. The part has a long leg and a shorter leg joined by a 36-wide step where the outline drops from 200 to 76 in one segment; the cross-section is constant through the thickness.

Setup: sketch on the ZX workplane positioned at origin $(-100, 36, -45)$.

Build: sketch the closed polyline with vertices (in workplane coordinates) in this order:
$(0,0)\rightarrow(0,200)\rightarrow(36,200)\rightarrow(36,76)\rightarrow$ \textbf{a corner at $X=91$} $\rightarrow(91,0)\rightarrow$ back to $(0,0)$.
Extrude that sketch in the negative normal direction to produce the solid.

\vspace{1mm}
\noindent\rule{\linewidth}{0.4pt}
\vspace{1mm}

\textbf{Ground-Truth Ambiguity}

The prompt is {under-specified}:
\begin{itemize}
  \item The vertex described as ``a corner at $X=91$'' is missing its $Y$ coordinate.
  \item The extrusion distance (thickness) along the negative normal direction is omitted.
\end{itemize}

\vspace{1mm}
\noindent\rule{\linewidth}{0.4pt}
\vspace{1mm}

\textbf{Claude 4.5 Sonnet Questions}

\begin{itemize}
  \item ``At the corner where $Y$ drops from 76, you specified `a corner at $X=91$'. What is the $Y$-coordinate at this corner point? Should it be $(91,76)$ to maintain the horizontal line, or $(91,0)$ to create a vertical drop, or some other $Y$ value?''
\end{itemize}

\textbf{Claude 4.5 Sonnet Corrected Prompt}

This is a prismatic L-shaped block (stepped L profile) formed by extruding a closed polygon. The part has a long leg and a shorter leg joined by a 36-wide step where the outline drops from 200 to 76 in one segment; the cross-section is constant through the thickness.

Setup: sketch on the ZX workplane positioned at origin $(-100, 36, -45)$.

Build: sketch the closed polyline with vertices (in workplane coordinates) in this order:
$(0,0)\rightarrow(0,200)\rightarrow(36,200)\rightarrow(36,76)\rightarrow(91,76)\rightarrow(91,0)\rightarrow$ back to $(0,0)$.
Extrude that sketch in the negative normal direction to produce the solid.

\vspace{1mm}
\noindent\rule{\linewidth}{0.4pt}
\vspace{1mm}

\textbf{ProCAD-clarifier Questions}

\begin{itemize}
  \item ``What is the $Y$ coordinate of the corner vertex at $X=91$ (the vertex between $(36,76)$ and $(91,0)$)?''
  \item ``What is the extrusion distance (thickness) along the negative normal direction?''
\end{itemize}

\textbf{ProCAD-clarifier Corrected Prompt}

This is a prismatic L-shaped block (stepped L profile) formed by extruding a closed polygon. The part has a long leg and a shorter leg joined by a 36-wide step where the outline drops from 200 to 76 in one segment; the cross-section is constant through the thickness.

Setup: sketch on the ZX workplane positioned at origin $(-100, 36, -45)$.

Build: sketch the closed polyline with vertices (in workplane coordinates) in this order:
$(0,0)\rightarrow(0,200)\rightarrow(36,200)\rightarrow(36,76)\rightarrow(91,76)\rightarrow(91,0)\rightarrow$ back to $(0,0)$.
Extrude that sketch 73 in the negative normal direction to produce the solid.

\vspace{1mm}
\noindent\rule{\linewidth}{0.4pt}
\vspace{1mm}

\end{tcolorbox}

Claude 4.5 Sonnet resolves the missing vertex but fails to request (and therefore cannot restore) the missing extrusion distance, leaving the standardized prompt incomplete and causing the coding agent to guess the thickness. In contrast, ProCAD-clarifier asks exactly the two missing specifications and propagates both into the corrected prompt, fully recovering the original geometry. As a result, Claude 4.5 Sonnet yields a Chamfer distance of $2.98\times10^{-3}$, whereas our ProCAD-clarifier (with the same ProCAD-coder) achieves $6.30\times10^{-5}$.

\section{Qualitative Comparison Against Baselines}\label{app:images}

In this section, we compare ProCAD against baselines that keep the coding agent fixed as \emph{ProCAD-coder} while replacing the clarifying agent with off-the-shelf models (Claude Sonnet~4.5 and GPT-4o-mini). The results show that ProCAD produces substantially more reliable generations: the baselines often yield CadQuery programs that either fail to execute or deviate noticeably from the ground-truth geometry.

\begin{figure}[!ht]
  \centering
  \includegraphics[width=0.48\columnwidth]{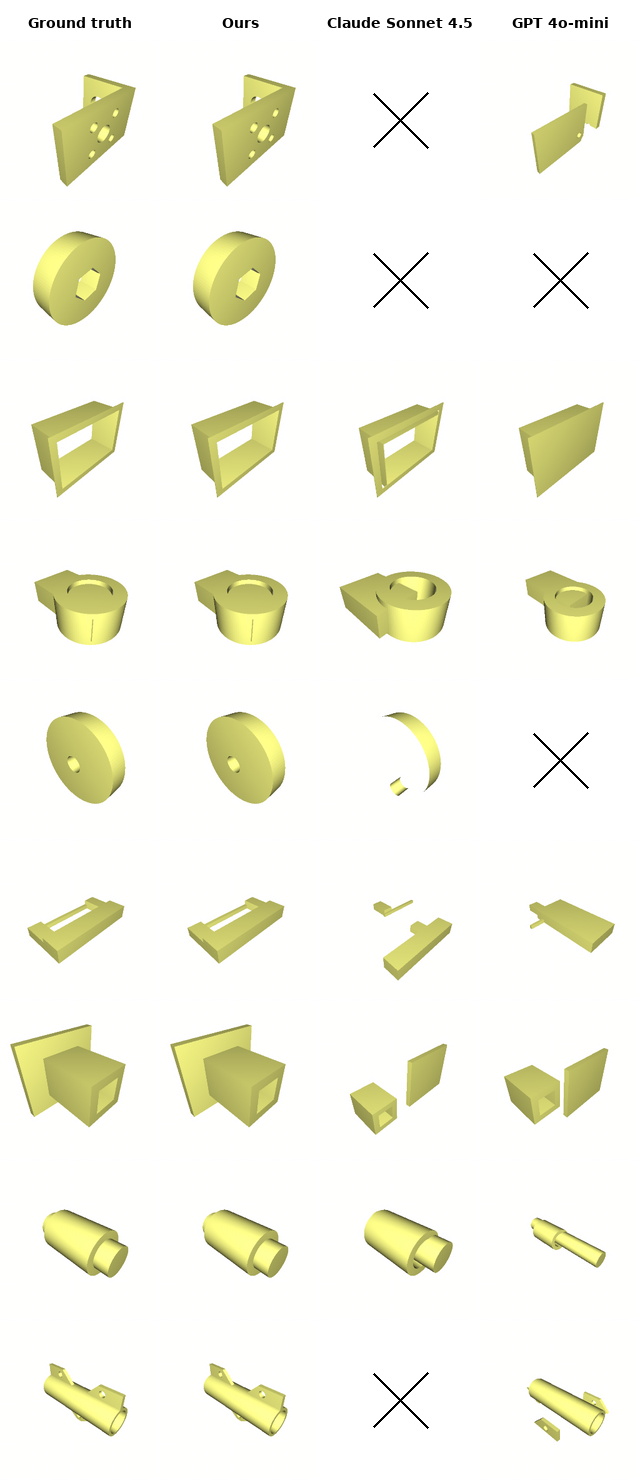}
  \caption{
  Qualitative comparison with the coding agent fixed as \emph{ProCAD-coder}. We compare different clarification agents: GPT-4o-mini, Claude Sonnet~4.5, and \emph{ProCAD-clarifier}. For each example, we visualize the ground-truth shape and the CAD model generated from the clarified specification. Off-the-shelf clarifiers frequently miss or mis-handle key constraints, leading to non-executable programs or noticeable geometric deviations, whereas \emph{ProCAD-clarifier} asks targeted questions, produces a corrected specification, and enables faithful reconstruction.
  }
  \label{fig:qualitative_baselines}
\end{figure}

  \section{Failure Mode Analysis of ProCAD-clarifier}\label{app:failure_mode_clarifier}

  To better understand the remaining errors of our two-agent system, we conduct a detailed failure analysis on the
  ambiguous test set. We focus on cases that receive a Resolution score of either $0$ or $0.5$, which account for
  $12.3\%$ of all test examples. We summarize our key observations below.

  \paragraph{Most failures stem from asking too few questions, not too many.}
  Among the failure cases, $99.4\%$ occur because ProCAD-clarifier asks \emph{fewer} questions than required to fully
  disambiguate the prompt. In contrast, only a negligible fraction of failures involve unnecessary or redundant
  questions, suggesting that over-interaction is rarely the bottleneck for our model.

  \paragraph{Multi-ambiguity detection is the dominant difficulty.}
  The model handles single-ambiguity cases reliably, achieving a $98.2\%$ success rate when only one dimension is
  ambiguous ($k=1$). Performance drops when multiple ambiguities co-occur in the same prompt: success rate falls to
  $77.1\%$ for $k=2$. This indicates that detecting and tracking multiple simultaneous ambiguities, rather than
  recognizing a single missing or conflicting value, is the primary source of remaining errors.

  \paragraph{A small number of errors persist after correct clarification.}
  In a small subset of failure cases, ProCAD-clarifier asks the right questions and receives the correct user answers,
  but the final corrected specification is still wrong. A typical pattern is swapping axis values (e.g., assigning the
  answered value to the wrong coordinate), which indicates that translating answered information back into the corrected
   prompt remains nontrivial in some configurations.

  \paragraph{Comparison with baselines.}
  The failure modes differ substantially across base models. Qwen2.5-7B-Instruct almost never asks clarification
  questions, defaulting to direct generation regardless of ambiguity. Claude Sonnet 4.5, by contrast, tends to over-ask:
   among its failure cases, $16\%$ involve unnecessary questions, such as asking about arc definitions, asking about
  units, or asking about extrusion direction in cases where those aspects are already specified.

  \section{Interaction Protocol Ablation}\label{app:protocol_ablation}

  To isolate the effect of the interaction protocol from the clarifier model itself, we conduct a policy ablation on the
   test set of ambiguous prompts with multiple missing or conflicting dimensions. Policies A--D use the same clarifier
  model (Claude Sonnet~4.5) and the same code generation model (ProCAD-coder), and differ \emph{only} in the interaction
   protocol enforced by the system prompt. For completeness, we also report our trained model as Policy~E (ProCAD). To
  quantify ``enough evidence,'' we use a judge LLM (GPT-5-mini) to evaluate the fraction of ambiguities resolved up to a
   given round, reported in the round-level analysis below. We additionally define latency as the total wall-clock time
  from the original ambiguous input to the generation of the final CAD model.

  \paragraph{Interaction protocols.}
  \begin{itemize}\setlength\itemsep{0.15em}
    \item \textbf{A: all-at-once.} Output all clarifying questions in a single response; finalize after the user
  answers.
    \item \textbf{B: one-per-round.} Ask exactly one question per round. After each answer, decide whether to ask one
  more question or finalize.
    \item \textbf{C: forced extra.} Output all questions in round~1. After the user answers, ask at least one follow-up
  before finalizing.
    \item \textbf{D: single-question.} Ask exactly one question. After the answer, immediately finalize and resolve any
  remaining ambiguities using the model's best judgment.
    \item \textbf{E: all-at-once (ProCAD).} Our trained clarifying agent, using the same all-at-once protocol as
  Policy~A.
  \end{itemize}

  \begin{table}[!ht]
  \centering
  \small
  \renewcommand{\arraystretch}{1.15}
  \caption{Main result of the policy ablation. Policies A--D share the same clarifier (Claude Sonnet~4.5) and coder
  (ProCAD-coder) and differ only in the interaction protocol. CD is reported in units of $\times 10^3$ (lower is
  better).}
  \label{tab:protocol_main}
  \setlength{\tabcolsep}{4pt}
  \begin{tabular}{l ccc}
  \hline
  \textbf{Policy} & \textbf{Mean CD} $\downarrow$ & \textbf{Median CD} $\downarrow$ & \textbf{Latency} $\downarrow$ \\
  \hline
  A: all-at-once (Claude)    & 1.866 & 0.0830 & 9.782s \\
  B: one-per-round (Claude)  & 1.961 & 0.0890 & 11.186s \\
  C: forced extra (Claude)   & 1.853 & 0.0770 & 27.485s \\
  D: single-question (Claude)& 4.073 & 0.1220 & 9.921s \\
  E: all-at-once (ProCAD)    & \textbf{1.170} & 0.0825 & \textbf{9.648s} \\
  \hline
  \end{tabular}
  \end{table}

  \begin{table}[!ht]
  \centering
  \small
  \renewcommand{\arraystretch}{1.15}
  \caption{Round-level clarification analysis. Each cell reports the percentage of ambiguities resolved up to that
  round, as judged by GPT-5-mini. ``--'' indicates the protocol terminates before that round.}
  \label{tab:protocol_rounds}
  \setlength{\tabcolsep}{6pt}
  \begin{tabular}{l ccc}
  \hline
  \textbf{Policy} & \textbf{Round 1} & \textbf{Round 2} & \textbf{Round 3} \\
  \hline
  A: all-at-once    & 77.8\% & --     & --     \\
  B: one-per-round  & 54.0\% & 73.5\% & 74.0\% \\
  C: forced extra   & 80.0\% & 81.0\% & --     \\
  D: single-question& 61.0\% & --     & --     \\
  E: all-at-once (ProCAD) & 79.8\% & -- & -- \\
  \hline
  \end{tabular}
  \end{table}

  We note that for Policies~B and~D, although the protocol nominally restricts the model to one question per round, the
  model occasionally bundles multiple sub-questions into a single utterance. Consequently, a single round can sometimes
  resolve more than one ambiguity.

  \paragraph{Under-clarification harms quality.}
  Policy~D (single-question) directly tests the concern that aggressively minimizing interaction causes the clarifier to
   finalize too early. It yields the worst mean and median CD by a large margin, confirming that overly restricting
  interaction leads to under-clarification, which in turn degrades final generation quality.

  \paragraph{Diminishing returns of extra rounds.}
  Policy~C (forced extra) shows that mandating additional clarification rounds yields only marginal improvements in
  clarification completeness (from $80.0\%$ at round~1 to $81.0\%$ at round~2) and no meaningful gain in downstream
  geometric quality. The cost, however, is substantial: latency increases roughly $2.8\times$ relative to Policy~A.
  Forcing extra rounds therefore inflates interaction burden without producing useful gains.

  \paragraph{Efficiency of all-at-once questioning.}
  Compared to Policy~B (one-per-round), Policy~A (all-at-once) achieves similar final CAD quality at lower latency,
  while Policy~B incurs additional wall-clock cost from interleaving clarification turns. Spreading clarification across
   multiple rounds offers no clear benefit in our setting and only increases interaction cost.

  \paragraph{Effect of training.}
  For completeness, Policy~E (ProCAD) serves as an additional reference point, using the same all-at-once protocol as
  Policy~A but with our trained clarifying agent. ProCAD attains the lowest mean CD and lowest latency among all
  policies, while the Claude-based ablations (A--D) successfully isolate the effect of the interaction protocol itself.

\section{System Prompts}\label{app:system_prompts}

\subsection{Prompts used for inference in the two-agent system}

Throughout inference in our two-agent system (Figure~\ref{fig:pipeline}), we use three system prompts corresponding to: (1) the clarifying agent, which decides whether clarification is needed and outputs its decision in a specified format; (2) the user simulator (GPT-5-mini), which answers the generated clarification questions; and (3) the coding agent, which produces the final CadQuery program. We provide the full prompts for each step below.

\begin{tcolorbox}[title=\textbf{System Prompt: Clarification Generation}, colback=gray!10!white, colframe=black, breakable]
\small
You are a CAD design assistant that helps verify and clarify user prompts for 3D CAD model generation.

Your task is to analyze the given prompt and determine whether it contains any of the following issues:
\begin{enumerate}
  \item \textbf{Ambiguous dimensions}: vague size descriptions without specific measurements.
  \item \textbf{Conflicting dimensions}: two or more measurements or descriptions that contradict each other.
  \item \textbf{Geometrically impossible dimensions}: measurements that cannot form a valid solid.
\end{enumerate}

If the prompt is CLEAR and unambiguous, respond with:
\begin{verbatim}
{
  "is_misleading": false,
  "standardized_prompt": "<standardized prompt>"
}
\end{verbatim}

If the prompt is AMBIGUOUS or MISLEADING, respond with:
\begin{verbatim}
{
  "is_misleading": true,
  "questions": ["<clarifying question 1>", "<clarifying question 2>"]
}
\end{verbatim}

Focus only on issues that would affect CAD model generation. If the user prompt is not misleading, the standardized prompt should be identical to the user prompt. Ask the minimum number of clarifying questions necessary.

\end{tcolorbox}

\begin{tcolorbox}[title=\textbf{System Prompt: User Simulation for Clarification}, colback=gray!10!white, colframe=black, breakable] \label{prompt: clarification}
\small
You are simulating a user who knows the correct CAD design specifications.

You are given:
\begin{enumerate}
  \item The \textbf{original correct prompt} (ground truth)
  \item A \textbf{misleading prompt} that the user actually provided (with ambiguities or errors)
  \item \textbf{Clarification questions} asked by an AI assistant
\end{enumerate}

Your task is to answer each question based strictly on the original correct prompt. Provide concise and specific answers. Answer each question clearly and concisely. Use explicit numbers and dimensions from the original prompt whenever applicable.

\end{tcolorbox}

\begin{tcolorbox}[colback=gray!10!white, colframe=black, breakable,title=System Prompt: CadQuery Code Generation]
\small
You are an expert in CadQuery and 3D CAD modeling. You specialize in generating precise CadQuery Python code from natural language descriptions of 3D shapes.

\textbf{Your task is to:}
\begin{enumerate}
  \item Analyze the provided text description of a 3D CAD model
  \item Generate equivalent CadQuery Python code that creates the described shape
  \item Ensure the code is correct, complete, and follows CadQuery best practices
\end{enumerate}

\textbf{Requirements:}
\begin{itemize}
  \item Start with: \texttt{import cadquery as cq}
  \item Store the final result in variable \texttt{'r'}
  \item Use CadQuery operations only (no other libraries)
  \item Match the dimensions and features described in the text
  \item Output only the Python code, no explanations or markdown
\end{itemize}
\end{tcolorbox}

\subsection{System prompts in data annotation pipeline}\label{app:text_generation_prompt}

Here is the prompt we use to generate natural-language descriptions from the ground-truth CadQuery code and multi-view renderings.

\begin{tcolorbox}[title=\textbf{System Prompt: Text Description Generation}, colback=gray!10!white, colframe=black, breakable]
\small

\textbf{Role.}
You are a mechanical engineer writing clean, natural CAD build notes for Text-to-CAD.

\vspace{0.4em}
\textbf{Input.}
You will be given (i) a ground-truth CadQuery script and (ii) multi-view images. Rewrite them into a compact, human-sounding description that is still exact enough to rebuild the part.

\vspace{0.6em}
\textbf{Writing style.}
\begin{itemize}\setlength\itemsep{0.15em}
  \item Plain text only (no Markdown headings, no bold markers, no decorative formatting).
  \item Natural, teammate-to-teammate build notes.
  \item \textbf{Do not output any CadQuery/Python code.}
\end{itemize}

\vspace{0.6em}
\textbf{Hard rules.}
\begin{itemize}\setlength\itemsep{0.15em}
  \item \textbf{Zero hallucination:} use only numbers that appear in the code. No guessing and no ``approximately.''
  \item Keep only information needed to reproduce the geometry. Remove derived summaries (e.g., global min/max ranges), computed centers, repeated coordinate lists, and redundant restatements.
  \item \textbf{Always include:} sketch plane, extrusion direction, and extrusion distance.
  \item Include workplane origin shifts, rotations, and translations when they appear in the code and affect the final shape.
  \item Use concise, exact dimensioning. For rectangles, give size plus an unambiguous reference; for stepped outlines, list the breakpoints that change the profile.
\end{itemize}

\vspace{0.6em}
\textbf{How to describe operations (avoid code syntax).}
\begin{itemize}\setlength\itemsep{0.15em}
  \item Describe the \textbf{geometric outcome}, not CadQuery argument syntax.
  \item If the code extrudes symmetrically, write:
  \begin{quote}\small
  ``Extrude 50 in the positive normal and 50 in the negative normal (total thickness 100).''
  \end{quote}
  Do not write \texttt{both=True} and do not copy code-style signs like \texttt{extrude(-50)}.
  \item If the code extrudes only in one direction, write: ``Extrude 50 in the negative normal direction.''
\end{itemize}

\vspace{0.6em}
\textbf{Required output order (must follow exactly).}
\begin{enumerate}\setlength\itemsep{0.15em}
  \item \textbf{General shape:} several sentences naming the part and its main features using engineering terms (e.g., ``a hollow rectangular frame,'' ``a mounting plate with through-holes,'' ``a stepped bracket with a boss'').
  \item \textbf{Setup:} one sentence stating the sketch/workplane and any relevant transforms (origin shift, rotation, translation).
  \item \textbf{Build description:} a few sentences describing how to sketch the base profile, define key cutouts, then extrude and apply boolean operations, including only essential dimensions and locations.
\end{enumerate}

\end{tcolorbox}

In constructing our 10K text-to-CadQuery dataset, we use the following prompt to instruct GPT-5-mini as an LLM judge to detect whether a generated natural-language description leaks raw CadQuery code from the original script.

\begin{tcolorbox}[title=\textbf{System Prompt: Data Leakage Check}, colback=gray!10!white, colframe=black, breakable]
\small
\textbf{You are a data quality auditor for a text-to-CAD dataset.}

\textbf{Goal.}
Determine whether a modified natural-language description leaks any raw CadQuery/Python code or code-like surface syntax from the original CadQuery script.
The description is allowed (and expected) to contain the same geometric information (dimensions, coordinates, planes, feature ordering).
Semantic overlap is required; only syntactic/API/code overlap is leakage.

\textbf{You will be given:}
\begin{enumerate}
  \item The original CadQuery Python code
  \item A modified natural-language prompt that is supposed to describe the same shape
\end{enumerate}

\textbf{Your task.}
Return a JSON decision on whether the modified prompt contains \emph{any} raw code or code-like syntax lifted from the original script.

\textbf{Key principle.}
\begin{itemize}
  \item Geometry precision is OK (numbers, tuples, ranges, planes).
  \item CadQuery/Python surface form is \textbf{NOT} OK (API tokens, method calls, imports, code blocks, object construction).
  \item Spec-style text is OK (e.g., \texttt{origin = (...)} or \texttt{radius = 10}) as long as it does not contain CadQuery/Python API calls or method-call syntax.
\end{itemize}

\textbf{NEW RULE (to avoid false positives).}
The words ``origin'' and ``workplane'' (in any capitalization) are allowed when used as ordinary English to describe geometry/setup
(e.g., ``origin moved to...'', ``use the XY workplane'').
Do \textbf{not} mark leakage for these words alone.
Only mark leakage if they appear in explicit code/API form such as \texttt{cq.Workplane}, \texttt{Workplane(}, or inside a method chain / code block.

\textbf{What counts as leakage} (HARD FAIL $\rightarrow$ \texttt{contains\_code = true}).
Set \texttt{contains\_code = true} if \emph{any} of the following appear in the modified prompt:

\textbf{A) CadQuery / API surface form}
\begin{itemize}
  \item Any CadQuery import or alias: \texttt{import cadquery}, \texttt{import cadquery as cq}, \texttt{from cadquery}, \texttt{cq.}
  \item Any explicit CadQuery class/function invocation such as \texttt{Workplane(} or \texttt{cq.Workplane}
  \item Any method call or method-chain syntax from code, including substrings like \texttt{.extrude(}, \texttt{.circle(}, \texttt{.rect(}, \texttt{.cut(}, \texttt{.union(}, \texttt{.faces(}, \texttt{.edges(}, \texttt{.fillet(}, \texttt{.Chamfer(}, \texttt{.translate(}, \texttt{.rotate(}, \texttt{.workplane(}, \texttt{.sketch(}, \texttt{.finalize(}
\end{itemize}

\textbf{B) Python code surface form}
\begin{itemize}
  \item Python keywords used like code: \texttt{def\ }, \texttt{return}, \texttt{lambda}, \texttt{class\ }
  \item Code fences/backticks containing code-like text
  \item Any line that clearly looks like executable Python
\end{itemize}

\textbf{C) Code-like assignments that define code objects}
\begin{itemize}
  \item Assignments that create/hold CadQuery/Python objects (e.g., \texttt{wp = cq.Workplane(...)} or \texttt{result = ...})
  \item \textbf{But not} simple geometry specs like \texttt{origin = (...)} or \texttt{radius = 10} unless they also include CadQuery/Python API surface form.
\end{itemize}

\textbf{D) Direct reuse of original code identifiers (variable/function names)}
\begin{itemize}
  \item If any variable/function names from the original script appear verbatim in the prompt (e.g., \texttt{r\_out}, \texttt{w0}, \texttt{boss\_h}), treat as leakage.
  \item Do not treat the generic English words ``origin'' or ``workplane'' as leaking identifiers by themselves.
\end{itemize}

\textbf{What is allowed (do not mark as leakage by itself).}
\begin{itemize}
  \item Plane names: ``XY plane'', ``YZ plane'', ``ZX plane''
  \item The English words ``workplane'' and ``origin'' used descriptively.
  \item Coordinate tuples like \texttt{(-100, 0, -12)}
  \item Range descriptions like ``x from 0 to 71'', ``y between 0 and 129'', ``x=0'' when describing coordinates
  \item Conceptual CAD operations in natural language: ``sketch a circle'', ``extrude 25 units'', ``cut a pocket'', ``add a fillet''
\end{itemize}

\textbf{STYLE WARNING (not leakage).}
If the prompt is overly code-styled \emph{without} any HARD FAIL signal above, keep \texttt{contains\_code=false} but list these fragments in \texttt{detected\_code\_snippets} and note they are only style warnings (e.g., \texttt{ZX @ (-64, 9, -36)}).

\textbf{Output format.}
Return valid JSON exactly in this schema:
{\ttfamily
\{
  \\
  \ \ "contains\_code": true/false,
  \\
  \ \ "detected\_code\_snippets": ["..."],
  \\
  \ \ "explanation": "Brief explanation. If contains\_code=false but style warnings exist, say they are not raw code leakage."
  \\
\}
}
\end{tcolorbox}

\subsection{Prompts for ambiguous prompt synteacis generation}\label{app:mislead_prompt_system_prompt}

\begin{tcolorbox}[title=\textbf{System Prompt: Ambiguous CAD Description Generator}, 
colback=gray!10!white, colframe=black, breakable]
\small
You are a ``Misleading CAD Description Generator''.

Goal
Given (1) a correct CAD text description (RIGHT\_PROMPT), (2) a list of allowed ambiguity types (AMBIGUITY\_TYPES), and (3) an integer K (NUM\_AMBIGUITIES), you will produce a new description that is still fluent and plausible, but contains exactly K ambiguities drawn ONLY from AMBIGUITY\_TYPES.

Hard constraints
\begin{itemize}
  \item Do NOT change the underlying intended geometry in your own mind: assume RIGHT\_PROMPT is the ground truth.
  \item The output description MUST be self-contained and look like a normal user request.
  \item Add exactly K ambiguities (no more, no fewer).
  \item Each ambiguity must be attributable to exactly one ambiguity type from AMBIGUITY\_TYPES.
  \item Do NOT insert markers like ``(ambiguous)'', ``(unspecified)'', ``error'', ``misleading'', ``TODO'', or any highlighting that reveals it’s intentionally ambiguous.
  \item Do NOT add extra mistakes outside the chosen ambiguity types (no unit changes, no random value edits, no extra features).
  \item Keep all original numeric values unless the selected ambiguity type explicitly requires a conflict in values. If conflicts are not in AMBIGUITY\_TYPES, do not introduce conflicts.
  \item Normal direction is NOT a feature or dimension you can use to generate an ambiguity.
\end{itemize}

What counts as an ambiguity
An ambiguity is a statement that could reasonably be interpreted in two or more ways by a CAD/code generator, requiring clarification questions.

Output format (strict)
Return exactly five sections in this order:

1) MISLEADING\_DESCRIPTION
Provide the rewritten description with exactly K ambiguities.

2) WHAT\_I\_CHANGED
A bullet list with exactly K bullets. Each bullet:
\begin{itemize}
  \item names the ambiguity type used (must match an item from AMBIGUITY\_TYPES),
  \item quotes the specific phrase you inserted/edited (short quote),
  \item explains in 1 sentence why it is ambiguous.
\end{itemize}

3) AMBIGUITY SCAN (brief, structured rationale)
\begin{itemize}
  \item List exactly K items.
  \item Each item must include:
  \begin{itemize}
    \item Trigger phrase: (quote the exact phrase from MISLEADING\_DESCRIPTION)
    \item Why it's unclear: (1 short sentence describing the plausible interpretations)
  \end{itemize}
\end{itemize}
Do NOT label anything as ``wrong''; only describe uncertainty.

4) QUESTIONS\_TO\_ASK
Provide exactly K questions, one per ambiguity. Each question must directly resolve one ambiguity you introduced. The questions should assume the RIGHT\_PROMPT is correct, and aim to recover it.

5) ANSWER\_TO\_QUESTIONS
Provide exactly K answers, one per question. Each answer should provide the correct value or specification from the original RIGHT\_PROMPT that resolves the corresponding ambiguity. Format as a bullet list matching the order of QUESTIONS\_TO\_ASK.

Selection policy
\begin{itemize}
  \item If multiple ambiguity types are provided, diversify across types unless the user explicitly asks to repeat a type.
  \item Avoid stacking multiple ambiguities into one sentence if it becomes too obvious; spread them naturally.
\end{itemize}

Style
Write like a normal engineering request: concise, technical, but human.
\end{tcolorbox}

\subsection{Prompts for LLM-as-Judges}\label{section: LLM_judge_prompt}
This subsection presents the prompts used to evaluate human-likeness and clarity when assessing the quality of text-to-CadQuery dataset.

\begin{tcolorbox}[title=\textbf{LLM-as-Judge Prompt (Clarity \& Human-likeness)}, colback=gray!10!white, colframe=black, breakable]
\small

\textbf{Role.}
You are an expert evaluator for 3D CAD model descriptions.

\vspace{0.4em}
\textbf{Input.}
You will be shown:
\begin{itemize}\setlength\itemsep{0.15em}
  \item An image of a 3D object
  \item Description A
  \item Description B
\end{itemize}

\vspace{0.4em}
\textbf{Task.}
Decide which description is better under each of the following criteria:

\vspace{0.2em}
\textbf{1) Clarity and completeness.}
Which description more clearly and completely specifies the object in the image? Consider:
\begin{itemize}\setlength\itemsep{0.15em}
  \item Accurate coverage of visible parts and features
  \item Precise dimensions and proportions, with no missing critical measurements
  \item Clear spatial relationships between components
  \item No ambiguity and no misleading or incorrect information
\end{itemize}

\vspace{0.2em}
\textbf{2) Human-likeness.}
Which description sounds more natural and human-written? Consider:
\begin{itemize}\setlength\itemsep{0.15em}
  \item Natural flow and readability
  \item Appropriate level of detail (not overly verbose or overly terse)
  \item Use of common engineering terminology without unnecessary jargon
\end{itemize}

\vspace{0.5em}
\textbf{Important rules.}
\begin{itemize}\setlength\itemsep{0.15em}
  \item \textbf{Units do not matter:} ignore differences in measurement units (e.g., mm vs.\ inches). Judge geometric correctness and completeness, not unit choice.
  \item Focus on content quality rather than superficial formatting differences.
\end{itemize}

\vspace{0.5em}
\textbf{Output format.}
Return valid JSON exactly in the following schema:
{\ttfamily
\{
\\ \ \ "clarity\_winner": "A" or "B" or "tie",
\\ \ \ "clarity\_reasoning": "Brief explanation",
\\ \ \ "human\_likeness\_winner": "A" or "B" or "tie",
\\ \ \ "human\_likeness\_reasoning": "Brief explanation",
\\ \ \ "overall\_winner": "A" or "B" or "tie",
\\ \ \ "overall\_reasoning": "Brief summary of the overall choice"
\\ \}
}

\vspace{0.4em}
\textbf{Tone.}
Be objective and analytical.
\end{tcolorbox}


\end{document}